\documentclass[10pt,twocolumn,letterpaper]{article}

\usepackage{cvpr}
\usepackage{times}
\usepackage{epsfig}
\usepackage{graphicx}
\usepackage{amsmath}
\usepackage{amssymb}
\usepackage{multirow}
\usepackage{booktabs}
\usepackage{subfig}
\usepackage{authblk}

\usepackage{enumitem}
\setitemize{noitemsep,topsep=0pt,parsep=0pt,partopsep=0pt}

\newcommand{\Sect}[1]{Section~\ref{#1}}
\newcommand{\Fig}[1]{Figure~\ref{#1}}
\newcommand{\Tbl}[1]{Table~\ref{#1}}
\newcommand{\Equ}[1]{Equation~(\ref{#1})}
\newcommand{\Alg}[1]{Algorithm~\ref{#1}}

\renewcommand{\paragraph}[1]{\textbf{#1}~~}

\usepackage[figuresright]{rotating}
\usepackage[ruled,vlined,algo2e]{algorithm2e}

\def\1{{\bf{1}}}
\def\0{{\bf{0}}}
\def\I{{\bf{I}}}
\def\w{{\bf w}}

\def\y{{\bf y}}

\def\a{{\bf a}}

\def\s{{\bf s}}
\def\rr{{\bf r}}

\def\cU{\mathcal{U}}

\def\cE{{\mathcal{E}}}
\def\cW{{\mathcal{W}}}
\def\cL{{\mathcal{L}}}

\newcommand{\R}{\mathbb{R}}

\newcommand{\E}{\mathbb{E}}

\DeclareMathOperator*{\argmin}{arg\,min}

\usepackage[breaklinks=true,bookmarks=false]{hyperref}
\cvprfinalcopy 

\def\cvprPaperID{5758} 

\ifcvprfinal\fi
\begin{document}


\title{ECC: Platform-Independent Energy-Constrained Deep Neural Network Compression via a Bilinear Regression Model}

\author[1]{Haichuan~Yang}
\author[1]{Yuhao~Zhu}
\author[2,1]{Ji~Liu}


\affil[1]{University of Rochester, Rochester, USA}
\affil[2]{Kwai Seattle AI Lab, Seattle, USA}
%

\maketitle
\thispagestyle{empty}

\begin{abstract}
Many DNN-enabled vision applications constantly operate under severe energy constraints such as unmanned aerial vehicles, Augmented Reality headsets, and smartphones. Designing DNNs that can meet a stringent energy budget is becoming increasingly important. This paper proposes ECC, a framework that compresses DNNs to meet a given energy constraint while minimizing accuracy loss. The key idea of ECC is to model the DNN energy consumption via a novel bilinear regression function. The energy estimate model allows us to formulate DNN compression as a constrained optimization that minimizes the DNN loss function over the energy constraint. The optimization problem, however, has nontrivial constraints. Therefore, existing deep learning solvers do not apply directly. We propose an optimization algorithm that combines the essence of the Alternating Direction Method of Multipliers (ADMM) framework with gradient-based learning algorithms. The algorithm decomposes the original constrained optimization into several subproblems that are solved iteratively and efficiently. ECC is also portable across different hardware platforms without requiring hardware knowledge. Experiments show that ECC achieves higher accuracy under the same or lower energy budget compared to state-of-the-art resource-constrained DNN compression techniques.
\end{abstract}

\section{Introduction}
\label{sec:intro}

Computer vision tasks are increasingly relying on deep neural networks (DNNs). DNNs have demonstrated superior results compared to classic methods that rely on hand-crafted features. However, neural networks are often several orders of magnitude more computation intensive than conventional methods~\cite{energygap, zhu2018algorithm}. As a result, DNN-based vision algorithms incur high latency and consume excessive energy, posing significant challenges to many latency-sensitive and energy-constrained scenarios in the real-world such as Augmented Reality (AR), autonomous drones, and mobile robots. For instance, running face detection continuously on a mobile AR device exhausts the battery in less than 45 minutes~\cite{likamwa2014draining}. Reducing the latency and energy consumption of DNN-based vision algorithms not only improves the user satisfaction of today's vision applications, but also fuels the next-generation vision applications that require ever higher resolution and frame rate.

Both the computer vision and hardware architecture communities have been actively engaged in improving the compute-efficiency of DNNs, of which a prominent technique is compression (e.g., pruning). Network compression removes unimportant network weights, and thus reduces the amount of arithmetic operations. However, prior work~\cite{yang2017designing, yang2018netadapt, yu2017scalpel} has shown that the number of non-zero weights in a network, or the network sparsity, does not directly correlate with execution latency and energy consumption. Thus, improving the network sparsity does not necessarily lead to latency and energy reduction.

Recognizing that sparsity is a poor, indirect metric for the actual metrics such as latency and energy consumption, lots of recent compression work has started directly optimizing for network latency~\cite{yang2018netadapt, he2018amc} and energy consumption~\cite{yang2017designing}, and achieve lower latency and/or energy consumption compare to the indirect approaches. Although different in algorithms and implementation details, these efforts share one common idea: they try to search the sparsity bound of each DNN layer in a way that the whole model satisfies the energy/latency constraint while minimizing the loss. In other words, they iteratively search the layer sparsity, layer by layer, until a given latency/energy goal is met. We refer to them as \textit{search-based} approaches.

The effectiveness of the search-based approaches rests on how close to optimal they can find the per-layer sparsity combination. Different methods differ in how they search for the optimal sparsity combination. For instance, NetAdapt~\cite{yang2018netadapt} uses a heuristic-driven search algorithm whereas AMC~\cite{he2018amc}\footnote{The method proposed in AMC originally targets model size, FLOPs or latency, but can be extended to target energy consumption using our modeling method introduced in~\Sect{sec:intro:model}.} uses reinforcement learning. However, the search-based approaches are fundamentally limited by the search space, which could be huge for deep networks.

In this paper, we propose an alternative DNN compression algorithm that compresses all the DNN layers together rather than compressing it layer by layer. This strategy eliminates many of the heuristics and fine-tuning required in previous layer-wise approaches. As a result, it is able to find compression strategies that lead to better latency and energy reductions. Due to the lack of compression techniques that specifically target energy consumption, this paper focuses on energy consumption as a particular direct metric to demonstrate the effectiveness of our approach, but we expect our approach to be generally applicable to other direct metrics as well such latency and model size.

The key to our algorithm is to use a differentiable model that numerically estimates the energy consumption of a network. Leveraging this model, we formulate DNN compression as a constrained optimization problem (constrained by a given energy budget). We propose an efficient optimization algorithm that combines ideas from both classic constrained optimizations and gradient-based DNN training. Crucially, our approach is \textit{platform-free} in that it treats the underlying hardware platform as a blackbox. Prior energy-constrained compressions all require deep understanding of the underlying hardware architecture~\cite{yang2017designing, yang2018end}, and thus are necessarily tied to a particular hardware platform of choice. In contrast, our framework directly measures the energy consumption of the target hardware platform without requiring any hardware domain knowledges, and thus is portable across different platforms.

\begin{figure}[t]
\vspace{-5pt}
  \centering
  \includegraphics[trim=0 0 0 0, clip, width=\columnwidth]{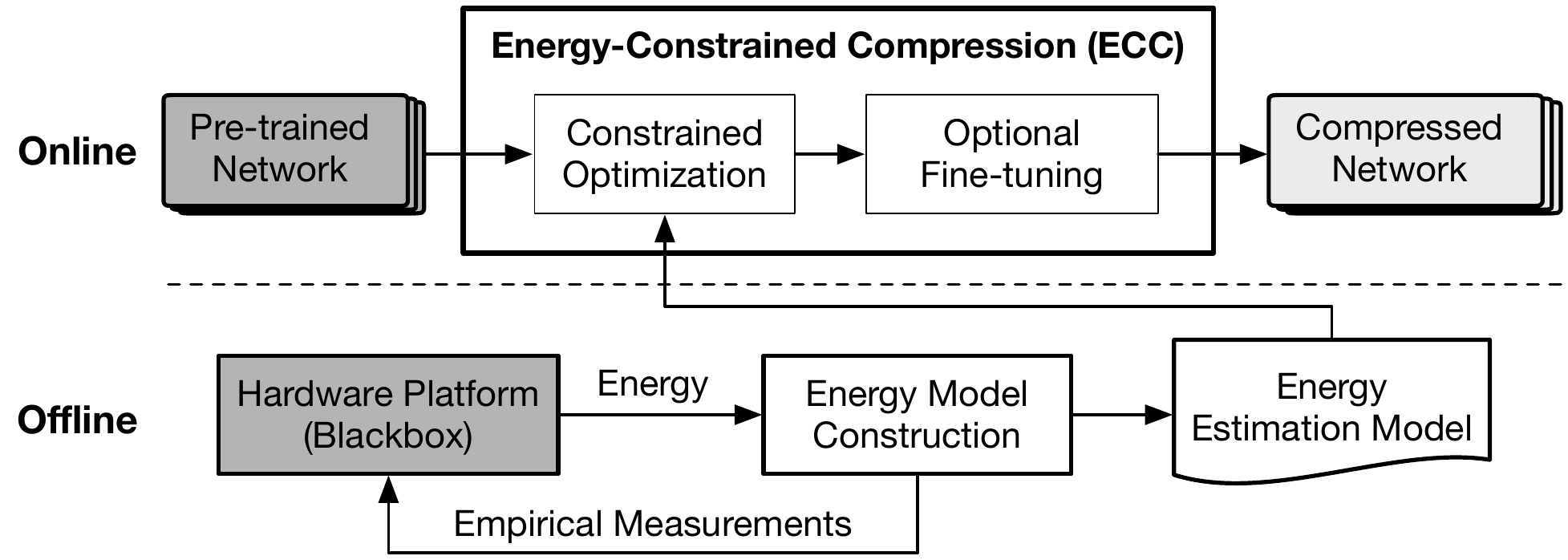}
  \caption{\small{ECC framework overview.}}
  \label{fig:framework}
  \vspace{-15pt}
\end{figure}

Leveraging the constrained optimization algorithm, we propose ECC, a DNN compression framework that automatically compresses a DNN to meet a given energy budget while maximizing its accuracy. ECC has two phases: an offline energy modeling phase and an online compression phase. Given a particular network to compress, the offline component profiles the network on a particular target platform and constructs an energy estimation model. The online component leverages the energy model to solve the constrained optimization problem followed by an optional fine-tuning phase before generating a compressed model. In summary, we make the following contributions:
\begin{itemize}[leftmargin=*]
	\item We propose a bilinear energy consumption model of DNN inference that models the DNN inference energy as a function of both its weights and sparsity settings. The energy model is constructed based on real hardware energy measurement and thus requires no domain-knowledge of the hardware architecture. Our model shows $<\pm 3\%$ error rate.
	\item We propose ECC, a DNN compression framework that maximizes the accuracy while meeting a given energy budget. ECC leverages the energy estimation model to formulate DNN compression as a constrained optimization problem. We present an efficient optimization algorithm that combines the classic ADMM framework with recent developments in gradient-based DNN algorithms. Although targeting energy in this paper, our framework can be extended to optimize other metrics such as latency and model size.
	\item We evaluate ECC using a wide range of computer vision tasks on both a mobile platform Jetson TX2 and a desktop platform with a GTX 1080 Ti GPU. We show that ECC achieves higher accuracy under the same energy budget compared to state-of-the-art resource-constrained compression methods including NetAdapt~\cite{yang2018netadapt} and AMC~\cite{he2018amc}.
\end{itemize}


\section{Related Work}
\label{sec:related}

\begin{table}[t]
	\Huge
	\centering
	\caption{\small Comparison across different   resource-constrained DNN compression techniques.}
	\renewcommand*{\arraystretch}{1.1}
	\renewcommand*{\tabcolsep}{10pt}
	\resizebox{\columnwidth}{!}
	{
		\begin{tabular}{l|cccc|c}
			\toprule[0.15em]
			Properties/Methods                              & EAP \cite{yang2017designing} & AMC \cite{he2018amc} & NetAdapt \cite{yang2018netadapt} & LcP \cite{chin2018layer} & \textbf{ECC} \\
			\midrule[0.05em]
			Use direct metric? & \checkmark & \checkmark & \checkmark & \checkmark & \checkmark \\
			Target energy? & \checkmark & & & & \checkmark \\
			Optimization-based? & & & & & \checkmark \\
			Platform-free? & & \checkmark & \checkmark & \checkmark & \checkmark \\
			\bottomrule[0.15em]
		\end{tabular}
	}
	\label{tab:comparison}
 \vspace{-15pt}
\end{table}


\paragraph{Network Compression} Network compression~\cite{lecun1990optimal} is a key technique in reducing DNN model complexity. It leverages the observation that some network weights have less impact on the final results and thus could be removed (zero-ed out). Compression techniques directly reduce a DNN model's size, which often also leads to latency and energy reductions. Early compression techniques focus exclusively on reducing the model size~\cite{han2015deep, han2015learning, wen2016learning, liu2015sparse, zhou2016less, li2016pruning} while latency and energy reductions are ``byproducts.'' It is well-known now that model size, latency, and energy consumption are not directly correlated~\cite{yang2017designing, yang2018netadapt, yu2017scalpel}. Therefore, compressing for one metric, such as model size, does not always translate to optimal compression results for other metrics such as latency and energy reduction, and vice versa.

\paragraph{Resource-Constrained Compression} Researchers recently started investigating resource-constrained compression, which compresses DNNs under explicit resource constraints (e.g., energy, latency, the number of Multiply-accumulate operations) instead of using model size as a proxy, though the model size could also be used as a constraint itself.~\Tbl{tab:comparison} compares four such state-of-the-art methods, EAP, AMC, and NetAdapt, and LcP. EAP~\cite{yang2017designing} compresses a model to reduce its energy consumption while meeting a given accuracy threshold. AMC~\cite{he2018amc} compresses a model to meet a given resource (size, FLOPs or latency) constraint while maximizing the accuracy. NetAdapt~\cite{yang2018netadapt} compresses a model to meet a given latency constraint while maximizing the accuracy.
LcP~\cite{chin2018layer} compresses a model to meet a given resource constraint (the number of parameters or the number of multiply-accumulate operations) while maximizing the accuracy.

Although the four techniques target different metrics and have different procedures, the core of them is to determine the optimal sparsity ratio of each layer in a way that the whole model meets the respective objectives. They differ in how they determine the layer-wise sparsity ratio. EAP, LcP, and NetAdapt all use heuristic-driven search algorithms. Specifically, EAP compresses layers in the order in which they contribute to the total energy, and prioritizes compressing the most energy-hungry layers; NetAdapt iteratively finds the per-layer sparsity by incrementally decreasing the latency budget; LcP assigns a score to each convolution filter, and prunes the filters based on the score until the resource constraint is satisfied. AMC uses reinforcement learning that determines the per-layer sparsity by ``trial and error.''

ECC is a different compression technique that, instead of compressing DNNs layer by layer, compresses all the network layers at the same time. It avoids heuristic searches and achieves better results.

\paragraph{Platform (In)dependence} Previous energy-constrained compressions~\cite{yang2017designing, yang2018end} rely on energy modeling that is tied to a specific hardware architecture. ECC, in contrast, constructs the energy model directly from real hardware measurements without requiring platform knowledges, and is thus generally applicable to different hardware platforms. AMC and NetAdapt are also platform-free as they take empirical measurements from the hardware, but they target latency and model size. In particular, NetAdapt constructs a latency model using a look-up table whereas ECC's energy model is differentiable, which is key to formulating DNN compression as a constrained optimization problem that can be solved using gradient-based algorithms.




\section{Method}
\label{sec:method}

This section introduces the proposed ECC framework for energy-constrained DNN compression. We first formulate DNN compression as a constrained optimization under the constraint of energy~(\Sect{sec:intro:prob}). We then describe how the energy is estimated using a bilinear model~(\Sect{sec:intro:model}). Finally, we explain our novel gradient-based algorithm that solves the optimization problem~(\Sect{sec:intro:opt}).

\subsection{Problem Formulation}
\label{sec:intro:prob}

Our objective is to minimize the loss function $\ell$ under a predefined energy constraint:
\begin{subequations}\label{eq:primal0}
	\begin{align}
	\min_{\cW} \quad & \ell(\cW)
	\\
	& \cE(\cW) \leq E_{\text{budget}},\label{eq:con}
	\end{align}
\end{subequations}
\noindent where $\cW := \{\w^{(u)}\}_{u \in \cU}$ ($\cU$ is the set of all layers) stacks the weights tensors of all the layers, and $\cE(\cW)$ denotes the real energy consumption of the network, which depends on the structure of DNN weights $\cW$. Compression affects the DNN weights $\cW$, and thus affects $\cE(\cW)$. $\ell$ is the loss function specific to a given learning task. In deep learning,  $\ell$ is a highly non-convex function.

There are two distinct classes of compression techniques. Unstructured, fine-grained compression prunes individual elements~\cite{han2015deep, han2015learning}; whereas structured, coarse-grained compression prunes a regular structure of a DNN such as a filter channel. While our method is applicable to both methods, we particularly focus on the coarse-grained method that prunes channels in DNN layers~\cite{srinivas2015data,zhou2016less,li2016pruning,liu2017learning,he2017channel,louizos2017bayesian,neklyudov2017structured,dai2018compressing} because channel pruning is more effective on off-the-shelf DNN hardware platforms such as GPUs~\cite{yu2017scalpel} whereas the fine-grained approaches require specialized hardware architectures to be effective~\cite{eyeriss, eie, yang2018end}.




With the channel pruning method, the optimization problem becomes finding the sparsity of each layer, i.e., the number of channels that are preserved in each layer, such that the total energy meets the given budget, that is,
\begin{subequations}
	\label{eq:primal}
	\begin{align}
	\min_{\cW, \s} \quad & \ell(\cW)
	\\
	\text{s.t.} \quad &\phi(\w^{(u)}) \leq s^{(u)}, \quad u \in \cU \label{eq:con_sparse1}
	\\
	& \cE(\s) \leq E_{\text{budget}}, \label{eq:con_energy}
	\end{align}
\end{subequations}
\noindent where $s^{(u)}$ corresponds to the sparsity bound of layer $u \in \cU$, and $\s:=\{s^{(u)}\}_{u \in \cU}$ stacks the (inverse) sparsities of all the layers. The energy consumption of a DNN $\cE$ can now be expressed as a function of $\s$\footnote{For simplicity, we reuse the same notion $\cE$ in both~\Equ{eq:con} and~\Equ{eq:con_energy}.}. $\w^{(u)}$ denotes the weight tensor of layer $u$. The shape of $\w^{(u)}$ is $d^{(u)} \times c^{(u)} \times r^{(u)}_h \times r^{(u)}_w$ for the convolution layer $u$ with $d^{(u)}$ output channels, $c^{(u)}$ input channels, and spatial kernel size $r^{(u)}_h \times r^{(u)}_w$. Without loss of generality, we treat the fully connected layer as a special convolution layer where $r^{(u)}_h = r^{(u)}_w = 1$. $\phi(\w^{(u)})$ calculates the layer-wise sparsity as $\sum_{i} \I(\|\w^{(u)}_{\cdot,i,\cdot,\cdot}\| \neq 0)$ where $\I(\cdot)$ is the indicator function which returns $1$ if the inside condition is satisfied and $0$ otherwise.

\subsection{Bilinear Energy Consumption Model}
\label{sec:intro:model}

The key step to solving~\Equ{eq:primal} is to identify the energy model $\cE(\s)$, i.e., to model the DNN energy as a function of the sparsity of each layer. This step is particularly important in that it provides an analytical form to characterize the energy consumption. Existing DNN energy models are specific to a particular hardware platform~\cite{yang2017designing, yang2018end}, which requires deep understandings of the hardware and is not portable across different hardware architectures. In contrast, we construct an energy model directly from hardware measurements while treating the hardware platform as a blackbox. This is similar in spirit to NetAdapt~\cite{yang2018netadapt}, which constructs a latency model through hardware measurements. However, their model is a look-up table that is huge and not differentiable. Our goal, however, is to construct a differentiable model so that the optimization problem can be solved using conventional gradient-based algorithms.


Our key idea is that the energy model can be obtained via a data driven approach. Let $\hat{\cE}$ be a differentiable function to approximate $\cE$:
 \vspace{-5pt}
\begin{equation}
\hat{\cE} = \argmin_{f\in \mathcal{F}} \E_{\s}[(f(\s) - \cE(\s))^2], \label{eq:e_pred}
 \vspace{-5pt}
\end{equation}

\noindent where $\mathcal{F}$ is the space of all the potential energy models, and $\E_{\s}$ is the expectation with respect to $\s:= [s_1,\cdots, s_{|\mathcal{U}|}, s_{|\mathcal{U}+1|}]$.

To find a differentiable energy model $\hat{\cE}$, our intuition is that the energy consumption of a DNN layer is affected by the number of channels in its input and output feature maps, which in turn are equivalent to the sparsity of the current and the next layer, respectively. Therefore, the energy consumption of layer $j$ can be captured by a function that models the interaction between $s_j$ and $s_{j+1}$, where $s_j$ denotes the (inverse) sparsity of layer $j$ ($j\in [1, |\cU|]$). Based on this intuition, we approximate the total network energy consumption using the following bilinear model:
$
\mathcal{F} := \{ f(\s) =a_0 + \sum_{j=1}^{|\cU|} a_j s_j s_{j+1}:~\ a_0, a_1,...,a_{|\cU|} \in \R_+\},
$
where $s_{|\cU|+1}$ is defined as the network output dimensionality, e.g., the number of classes in a classification task. Coefficients $a_0, a_1, ... a_{|\cU|}$ are the variables defining this space.

The rationale behind using the bilinear structure is that the total number of arithmetic operations (multiplications and additions) during a DNN inference with layers defined by $\s$ would roughly be in a bilinear form. Although other more complex models are possible, the bilinear model is simple and tight, which is easy to train and also avoids overfitting effectively. We will show in~\Sect{sec:eval} that our model achieves high prediction accuracy.

Sharp readers might ask what if the energy function fundamentally can not be modeled in the bilinear form (e.g., for some particular DNN architectures). In such a case, one could use a neural network to approximate the energy function since a three-layer neural network can theoretically approximate any function~\cite{hornik1991approximation}. Note that our constrained optimization formulation requires only that the energy model is differentiable and thus is still applicable.

To obtain $\hat{\cE}$, we sample $\s$ from a uniform distribution and measure the real energy consumption on the target hardware platform to get $\cE(\s)$. Please refer to~\Sect{sec:eval:setup} for a complete experimental setup. We then use the stochastic gradient descent (SGD) method to solve~\Equ{eq:e_pred} and obtain $\hat{\cE}$. Note that this process is performed once for a given network and hardware combination. It is performed offline as shown in~\Fig{fig:framework} and thus has no runtime overhead.  

\subsection{Optimization Algorithm}
\label{sec:intro:opt}

The optimization problem posed by~\Equ{eq:primal} is a constrained optimization whereas conventional DNN training is unconstrained. Therefore, conventional gradient-based algorithms such as SGD and existing deep learning solvers do not directly apply here. Another idea is to extend the gradient-based algorithm to the projected version \cite{shamir2013stochastic} -- a (stochastic) gradient descent step followed by a projection step to the constraint~\cite{yang2018end}. When the projection is tractable~\cite{ye2018unified} or the constraint is simple (e.g., linear)~\cite{pathak2015constrained}, Lagrangian methods can be used to solve constraints on the parameters or the outputs of DNNs. However, due to the complexity of our constraint, the projection step is extremely difficult.

In this paper, we apply the framework of ADMM~\cite{boyd2011distributed}, which is known to handle constrained optimizations effectively. ADMM is originally designed to solve optimization problems with linear equality constraints and convex objectives, both of which do not hold in our problem.
Therefore, we propose a hybrid solver that is based on the ADMM framework while taking advantage of the recent advancements in gradient-based deep learning algorithms and solvers~\cite{kingma2014adam}.

\paragraph{Algorithm Overview} We first convert the original problem \eqref{eq:primal} to an \emph{equivalent} minimax problem~\cite{boyd2011distributed}:
 \vspace{-5pt}
\begin{equation}
\min_{\cW, \s} \max_{z\geq 0, \y\geq \0} \cL (\cW, \s, \y, z)  \label{eq:minmax}
 \vspace{-5pt}
\end{equation}
\noindent where $\y$ is the dual variable introduced for the constraint~\eqref{eq:con_sparse1}, and $z$ is the dual variable for the constraint~\eqref{eq:con_energy}. $\cL$ is defined as the augmented Lagrangian $\cL(\cW, \s, \y, z) := \ell(\cW) + \cL_1(\cW, \s, \y) +\cL_2(\s, z)$, where
$
\cL_1(\cW, \s, \y) :=  {\rho_1 \over 2} \sum_u [\phi(\w^{(u)}) - s^{(u)} ]_+^2 
+ \sum_u y^{(u)} (\phi(\w^{(u)}) - s^{(u)}), 
\cL_2(\s, z) := {\rho_2 \over 2} [\hat{\cE}(\s) - E_{\text{budget}}]_+^2 + z (\hat{\cE}(\s) - E_{\text{budget}}).
$
$[\cdot]_+$ denotes the nonnegative clamp $\max(0, \cdot)$, and $\rho_1, \rho_2$ are two predefined nonnegative hyper-parameters. Note that the choices of $\rho_1$ and $\rho_2$ affect only the efficiency of convergence, but not the convergent point (for convex optimization).

To compress a DNN under a given energy constraint, we start with a dense model denoted by $\cW^{\text{dense}}$ (which could be obtained by optimizing the unconstrained objective), and solve the problem in~\Equ{eq:minmax} to obtain a compressed network. Inspired by the basic framework of ADMM, we solve~\Equ{eq:minmax} iteratively, where each iteration updates the primal variables $\cW, \s$ and dual variables $\y, z$. \Alg{alg:admm} shows the pseudo-code of our algorithm.

\begin{algorithm2e}[t]
	\SetAlgoLined
	\KwIn{Energy budget $E_{\text{\rm budget}}$, learning rates $\alpha, \beta$, penalty parameters $\rho_1, \rho_2$.}
	\KwResult{DNN weights $\cW^{*}$.}
	Initialize $\cW = \cW^{\text{dense}}$, $\s = \{\phi(\w^{(u)})\}_{u\in\cU}$, $\y=\0$, $z=0$;\\
	\While{$\hat{\cE}(\s) > E_{\text{budget}}$ or $\exists u, \phi(\w^{(u)}) > s^{(u)}$}{
		Update $\cW$ by proximal Adam update: $\cW$ = \eqref{eq:minw2};\\
		Update $\s$ by gradient descent:~\eqref{eq:mins};\\
		Update $\y, z$ by projected gradient ascent: \eqref{eq:maxy} and \eqref{eq:maxz};\\
	}
	$\cW^* = \cW$.
	\caption{Energy-Constrained DNN Compression.}
	\label{alg:admm}
\end{algorithm2e}

Specifically, each iteration first updates the DNN weights $\cW$ to minimize $\ell$ while preventing $\cW$ to have layer-wise (inverse) sparsities larger than $\s$. $\s$ is then updated to reduce the energy estimation $\hat{\cE}(\s)$. Dual variables $\y$ and $z$ can be seen as penalties that are dynamically changed based on how much $\cW$ and $\s$ violate the constraints~\eqref{eq:con_sparse1} and ~\eqref{eq:con_energy}. We now elaborate on the three key updating steps.


\subsubsection{Updating Primal Variable $\cW$}
\label{sec:method:updatew}

We first fix the sparsity bounds $\s$ and the two dual variables $\y, z$ to update the primal variable weight tensor $\cW$ by:
 \vspace{-5pt}
\begin{equation}
\argmin_{\cW} \ell(\cW) + \cL_1(\cW, \s, \y). \label{eq:minw}
 \vspace{-5pt}
\end{equation}

The challenge here is that updating $\cW$ is time-consuming, mainly due to the complexity of calculating $\argmin_{\cW} \ell(\cW)$, where $\ell$ is the non-convex loss function of the network. Stochastic ADMM~\cite{zheng2016fast} could simplify the complexity of the primal update by using stochastic gradient, but they consider only the convex problems with shallow models. Instead, we propose to improve the primal update's efficiency using a proxy of the loss $\ell(\cW)$ at $\mathcal{W}^t$:
\begin{equation}
\ell(\cW^t) + \langle \hat{\nabla}\ell(\cW^t), \cW - \cW^t\rangle + {1\over 2 \alpha} \|\cW - \cW^t\|_B^2,
\end{equation}
\noindent where $B$ is a positive diagonal matrix, which is usually used in many adaptive optimizers such as ADADELTA~\cite{zeiler2012adadelta} and Adam~\cite{kingma2014adam}; $\|\cW\|_{B}$ is defined by the norm $\sqrt{\text{vec}(\cW)^\top B\text{vec}(\cW)}$ where $\text{vec}(\cdot)$ is the vectorization operation. Without loss of generality, we use the diagonal matrix $B$ as in Adam. $\hat{\nabla}\ell(\cW^t)$ is the stochastic gradient of $\ell$ at $\cW^t$, and $\alpha$ is the learning rate for updating $\cW$. Therefore, \Equ{eq:minw} is simplified to a proximal~\cite{parikh2014proximal} Adam update:
 \vspace{-5pt}
\begin{equation}
\argmin_{\cW} {1\over 2\alpha} \| \cW - (\cW^{t} - \alpha{B^{-1}}\hat{\nabla} \ell(\cW^{t})) \|^2_{B} + \cL_1(\cW, \s, \y). \label{eq:minw2}
\end{equation}

If we define $\textbf{prox}_{\alpha\cL_1}(\cdot)$ as the proximal operator of function $\alpha\cL_1(\cdot, \s, y)$:
 \vspace{-5pt}
\begin{equation}
\textbf{prox}_{\alpha\cL_1}(\bar{\cW}) := \argmin_{\cW} {1\over 2} \| \cW - \bar{\cW}\|^2_{B} + \alpha\cL_1(\cW, \s, \y),
\end{equation}

\noindent the optimal solution of problem~\eqref{eq:minw2} admits a closed form: $\textbf{prox}_{\alpha\cL_1}(\cW^{t} - \alpha{B^{-1}}\hat{\nabla} \ell(\cW^{t}))$. This update essentially performs pruning and fine-tuning simultaneously. The detailed algorithm for proximal operator $\textbf{prox}_{\alpha\cL_1}(\cdot)$ is shown in \Alg{alg:prox}.


\begin{algorithm2e}[t]
	\SetAlgoLined
	\KwIn{Input tensors $\bar{\cW}=\{\bar{\w}^{(u)}\}_{u\in \cU}$.}
	\KwResult{Proximal operation result $\cW=\{{\w}^{(u)}\}_{u\in \cU}$.}
	Let $\a^{(u)}_i = \| \bar{\w}^{(u)}_{\cdot,i,\cdot,\cdot}\|^2_{B^{(u)}}, \forall u \in \cU$;\\
	Sort $\a^{(u)}$ in descending order, let $\rr^{(u)}$ be the corresponding ranks of elements in $\a^{(u)}$;\\
	\ForEach{Layer $u \in \cU$}{
		\For{$i \leftarrow 1$ \KwTo $c^{(u)}$}{
			\uIf{$\a^{(u)}_i > {\rho_1 \alpha}([\rr^{(u)}_i -  s^{(u)}]_+^2 - [\rr^{(u)}_i - 1 - s^{(u)}]_+^2 ) + {2\alpha}y^{(u)}$}{
				$\w^{(u)}_{\cdot,i,\cdot,\cdot} =  \bar{\w}^{(u)}_{\cdot,i,\cdot,\cdot}$;
			}
			\Else{
				$\w^{(u)}_{\cdot,i,\cdot,\cdot} = \0$;
			}
		}
	}
	\caption{Proximal Operator $\textbf{prox}_{\alpha\cL_1}(\cdot)$.}
	\label{alg:prox}
\end{algorithm2e}
\vspace{-5pt}

\subsubsection{Updating Primal Variable $\s$}
\label{sec:method:updates}

In this step, we update the primal variable $\s$ by:
 \vspace{-5pt}
\begin{equation}
\argmin_\s \cL_1(\cW, \s, \y) +\cL_2(\s, z).
 \vspace{-5pt}
\end{equation}

Similar as above, instead of searching for exactly solving this subproblem, we only apply a gradient descent step:
\begin{equation}
\s^{t+1} = \s^t - \beta (\nabla_\s \cL_1(\cW, \s^t, \y) +\nabla_\s \cL_2(\s^t, z)),\label{eq:mins}
\end{equation}
where $\beta$ is the learning rate for updating $\s$. To avoid removing a certain layer entirely, a lower bound is set for $s^{(u)}$. In our method, it is set as 1 if not explicitly mentioned.

\subsubsection{Updating Dual Variables}
\label{sec:method:updateyz}

The dual updates simply fix $\cW, \s$ and update $\y, z$ by projected gradient ascent with learning rates $\rho_1, \rho_2$:
\begin{align}
{y^{(u)}}^{t+1} &= [{y^{(u)}}^{t} + \rho_1(\phi(\w^{(u)}) - s^{(u)})]_+, \label{eq:maxy}\\
z^{t+1} &= [z^{t} + \rho_2 (\hat{\cE}(\s) - E_{\text{budget}})]_+. \label{eq:maxz}
\end{align}
To stabilize the training process, we perform some additional steps when updating the dual variables.
The dual variable $\y$ controls the sparsity of each DNN layer, and larger $y^{(u)}$ prunes more channels in layer $u$. It is not necessary to penalize $\phi(\w^{(u)})$ when $\phi(\w^{(u)}) \leq s^{(u)}$, so $y^{(u)}$ is trimmed to meet $\phi(\w^{(u)}) \geq \lfloor s^{(u)} \rfloor$. The dual variable $z$ is used to penalize the violation of energy cost $\hat{\cE}(\s)$, and larger $z$ makes $\s$ prone to decrease $\hat{\cE}(\s)$. In the training process, we want $\hat{\cE}(\s)$ to be monotonically decreased. So we project the variable $z$ to be large enough to meet $\max(\nabla_\s \cL_1(\cW, \s, \y) +\nabla_\s \cL_2(\s, z)) \geq \epsilon$, where $\epsilon$ is a small positive quantity and we simply set $10^{-3}$. The gradient of $\s$ is also clamped to be nonnegative.

\section{Evaluation Results}
\label{sec:eval}

We evaluate ECC on real vision tasks deployed on two different hardware platforms. We first introduce our experimental setup~(\Sect{sec:eval:setup}), followed up by the accuracy of the energy prediction model~(\Sect{sec:eval:epred}). Finally, we compare ECC with state-of-the-art methods~(\Sect{sec:eval:res}).

\subsection{Experimental Setup}
\label{sec:eval:setup}

\paragraph{Vision tasks \& Datasets} We evaluate ECC on two important vision tasks: image classification and semantic segmentation. For image classification, we use the complete ImageNet dataset~\cite{ILSVRC15}. For semantic segmentation, we use the recently released large-scale segmentation benchmark Cityscapes~\cite{Cordts2016Cityscapes} which contains pixel-level high resolution video sequences from 50 different cities.

\paragraph{DNN architectures} For image classification, we use two representative DNN architectures AlexNet~\cite{krizhevsky2012imagenet} and MobileNet~\cite{howard2017mobilenets}. The dense versions of the two models are from the official PyTorch model zoo and the official TensorFlow repository~\cite{abadi2016tensorflow}, respectively. For semantic segmentation, we choose the recently proposed ERFNet~\cite{romera2018erfnet}, which relies on residual connections and factorization structures, and is shown to be efficient on real-time segmentation. We use the pre-trained ERFNet released by the authors. The collection of the three networks allows us to evaluate ECC against different DNN layer characteristics including fully connected, convolutional, and transposed convolutional layers.

\paragraph{Hardware Platforms} We experiment on two different GPU platforms. The first one is a GTX 1080~Ti GPU. We use the \texttt{nvidia-smi} utility~\cite{nvidiasmi} to obtain real-hardware energy measurements. The second one is the Nvidia Jetson TX2 embedded device, which is widely used in mobile vision systems and contains a mobile Pascal GPU. We retrieve the TX2's GPU power using the Texas Instruments INA 3221 voltage monitor IC through the I2C interface. The DNN architectures as well as our ECC framework are implemented using PyTorch~\cite{paszke2017automatic}.

\paragraph{Baseline} We compare ECC with two most recent (as of submission) resource-constrained compression methods NetAdapt~\cite{yang2018netadapt} and AMC~\cite{he2018amc}. We faithfully implement them according to what is disclosed in the papers. NetAdapt is originally designed to compress DNNs under latency constraints and AMC is designed to compress DNNs under constraint of model size, FLOPs or latency. We adapt them to obtain the energy-constrained versions for comparison. Both methods use channel pruning for compression, same as ECC.

Earlier channel pruning methods such as Network-Slimming~\cite{liu2017learning}, Bayesian Compression~\cite{louizos2017bayesian}, and several others~\cite{dai2018compressing, he2017channel, neklyudov2017structured} are agnostic to resource constraints (e.g., energy) because they focus on sparsity itself. They require a sparsity bound (or regularization weight) for each layer to be \textit{manually} set before compression. The compressed model is generated only based on the given sparsity bounds, regardless of the energy budget. We thus do not compare with them here.


\paragraph{Hyper-parameters \& implementation details} The batch size is set to 128 for AlexNet and MobileNet and to 4 for ERFNet based on the GPU memory capacity. We use the Adam optimizer with its default Beta $(0.9,0.999)$; its learning rate is set to $10^{-5}$, and the weight decay is set as $10^{-4}$.\footnote{They are chosen in favor of best pre-trained model accuracy rather than biasing toward any compression methods.} All the compression methods are trained with the same number of data batches / iterations, which are about 300,000 for ImageNet and 30,000 for Cityscapes. These iterations correspond to the ``short-term fine-tuning''~\cite{yang2018netadapt} in NetAdapt and the ``4-iteration pruning \& fine-tuning''~\cite{he2018amc} in AMC. The reinforcement learning episodes in AMC is set to 400 as described in~\cite{he2018amc}. For the loss function $\ell$, we add a knowledge distillation (KD) term~\cite{ba2014deep} combined with the original loss (e.g., cross-entropy loss), since KD has been shown as effective in DNN compression tasks~\cite{mishra2017apprentice, tschannen2017strassennets, yang2018end}. In our method, the learning rate $\beta$ for the sparsity bounds $\s$ is set to reach the energy budgets with given iteration number, and the dual learning rates $\rho_1,\rho_2$ are set as 10 on ImageNet and 1 on Cityscapes.

After getting the compressed models with given energy budgets, we fine-tune each model for 100,000 iterations with aforementioned setup. For MobileNet, we additionally perform 300,000 iterations of fine-tuning with decayed learning rate (cosine decay from $3\times 10^{-5}$) to minimize the cross-entropy loss. The fine-tuning procedures train the DNN with fixed non-zero positions in their weight tensors.

\subsection{Energy Prediction Model}
\label{sec:eval:epred}

To train the energy prediction model, we obtain the real energy measurements under different layer-wise sparsity bounds $\s$. We first randomly sample $\s$ from the uniform distribution: $s^{(u)} \sim \text{unif}\{1, c^{(u)}\}$. For each sample, we then construct a corresponding DNN, and measure its energy $\cE({\s})$. We measure the energy by taking the average of multiple trials to minimize offset run-to-run variation. For each DNN architecture, we collect 10,000 $(\s, \cE(\s))$ pairs to train the energy model $\hat{\cE}$. We randomly choose 8,000 pairs as the training data and leave the rest as test data. To optimize problem~\eqref{eq:e_pred}, we use Adam optimizer with its default hyper-parameters, and the weight decay is set as 1.0. We set batch size as 8,000 (full training data) and train the energy model with 10,000 iterations. It should be noted that the bilinear energy model is linear in terms of the learnable parameters, which means a linear regression solver could be used. In this section, we also compare the bilinear model with a nonlinear model, for which Adam is more suitable.

\begin{figure}[t]
	\centering
	\subfloat[\small{MobileNet on GTX 1080~Ti.}]
	{
		\includegraphics[width=0.48\linewidth]{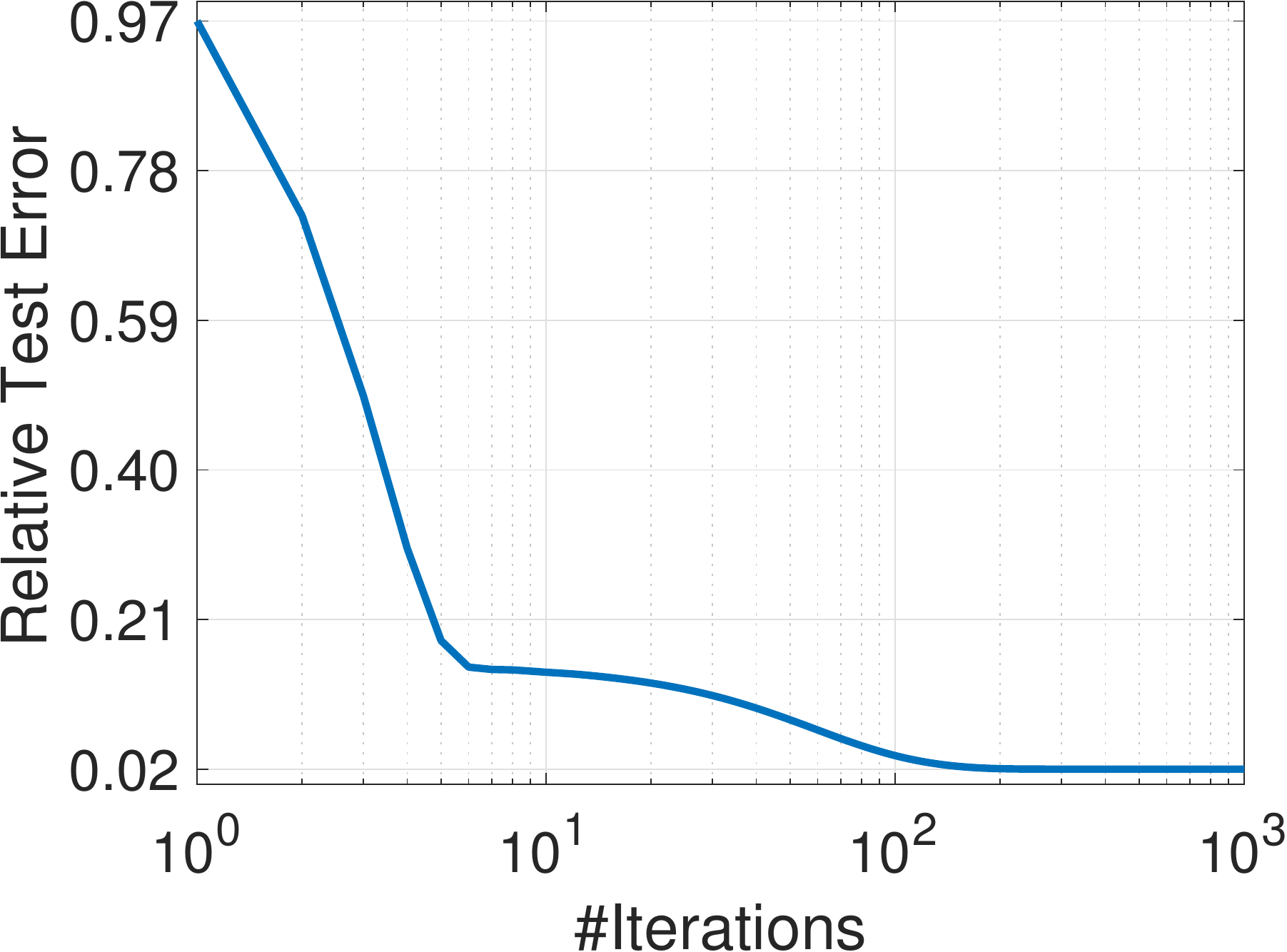}
	}
	\subfloat[\small{MobileNet on TX2.}]
	{
		\includegraphics[width=0.48\linewidth]{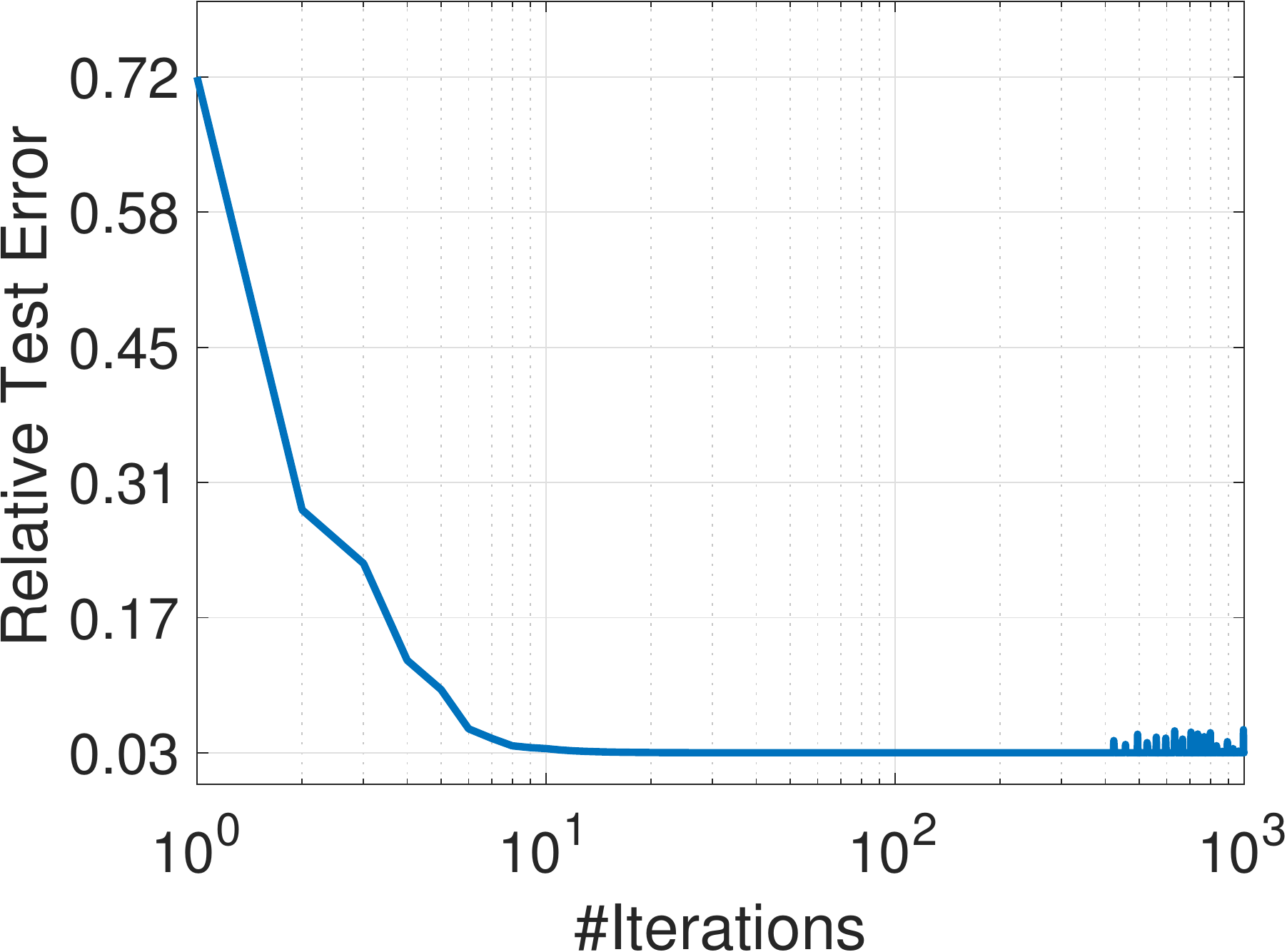}
	}
	\vspace{-5pt}
	\caption{Relative test error of energy prediction using the proposed bilinear model.}
	\label{fig:energy_mb}
	\vspace{-10pt}
\end{figure}

In \Fig{fig:energy_mb}, we show the relative test error defined as
$
\E_{\s\sim \text{testset}} [{|\hat{\cE}(\s) - \cE(\s)| /  \cE(\s)}]
$
at each training iteration for MobileNet on both hardware platforms. We find that the relative test errors quickly converge to around 0.03. This indicates that our energy model is not only accurate, but is also efficient to construct. The same conclusions hold for other networks as well, but are omitted due to space limit.

\begin{figure}[t]
	\centering
	\subfloat[\small{MobileNet on GTX 1080~Ti.}]
	{
		\includegraphics[width=0.48\linewidth]{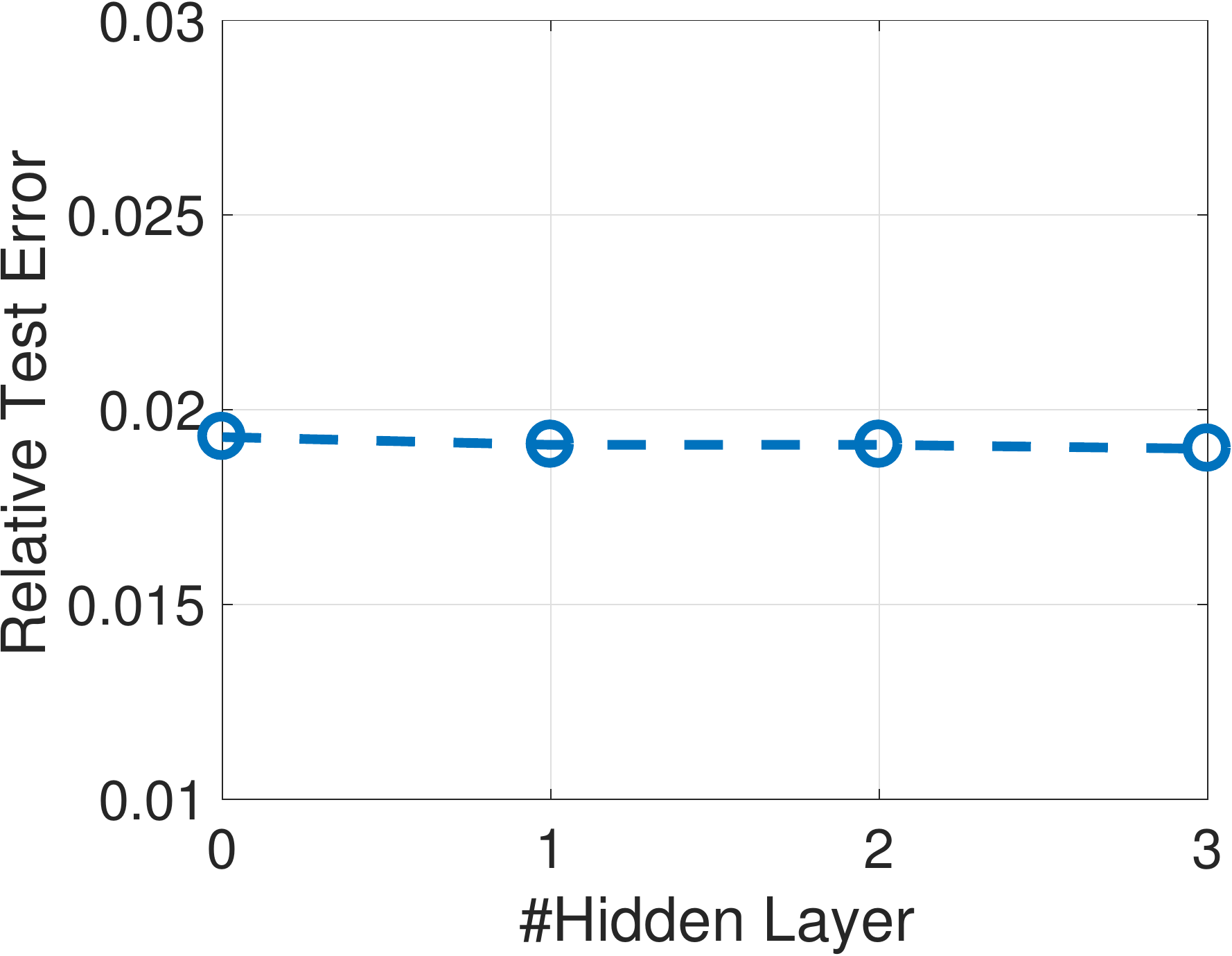}
	}
	\subfloat[\small{MobileNet on TX2.}]
	{
		\includegraphics[width=0.48\linewidth]{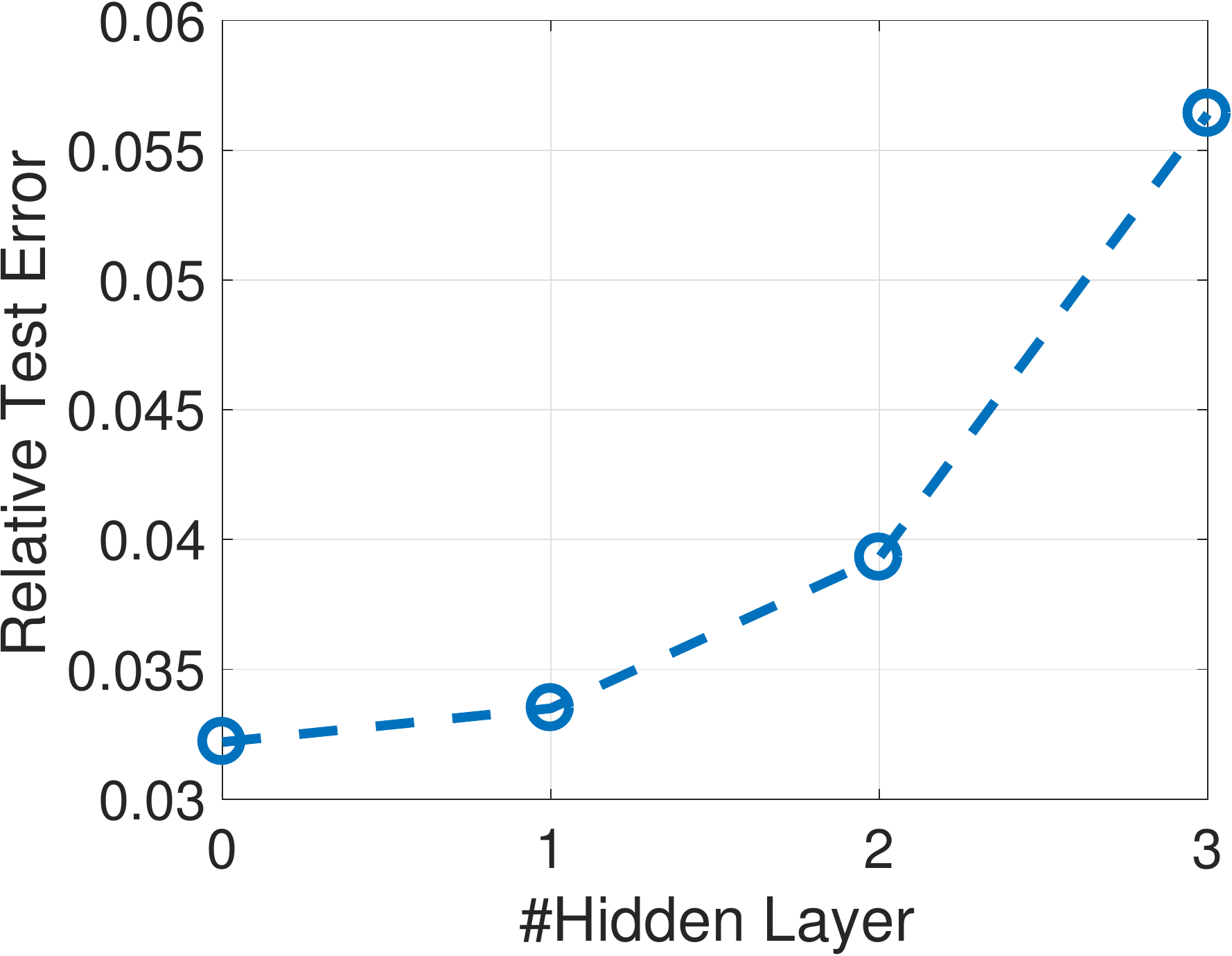}
	}
	\vspace{-5pt}
	\caption{Relative test error of energy prediction using an MLP model with different hidden layers.}
	\label{fig:edep_mb}
\end{figure}

To assess whether the bilinear model is sufficient, we also experiment with a more complex prediction model by appending a multilayer perceptron (MLP) after the bilinear model. The widths of all the MLP hidden layers  are set to 128, and we use the SELU~\cite{klambauer2017self} activation. We vary the number of hidden layers from 1 through 3. The prediction errors of this augmented energy model on MobileNet are shown in~\Fig{fig:edep_mb}, where the original bilinear model has zero hidden layer. We find that adding an MLP does not noticeable improve the prediction accuracy on GTX 1080~Ti, but significantly \textit{reduce} the prediction accuracy on TX2. We thus use the plain bilinear model for the rest of the evaluation.

\subsection{DNN Compression Results}
\label{sec:eval:res}

We now show the evaluation results on two popular vision tasks: image classification and semantic segmentation.
\subsubsection{ImageNet Classification}
\label{sec:eval:res:imagenet}

\begin{figure}[htbp]
	\centering
	\subfloat[\small{GTX 1080~Ti.}]
	{
		\includegraphics[width=0.48\columnwidth]{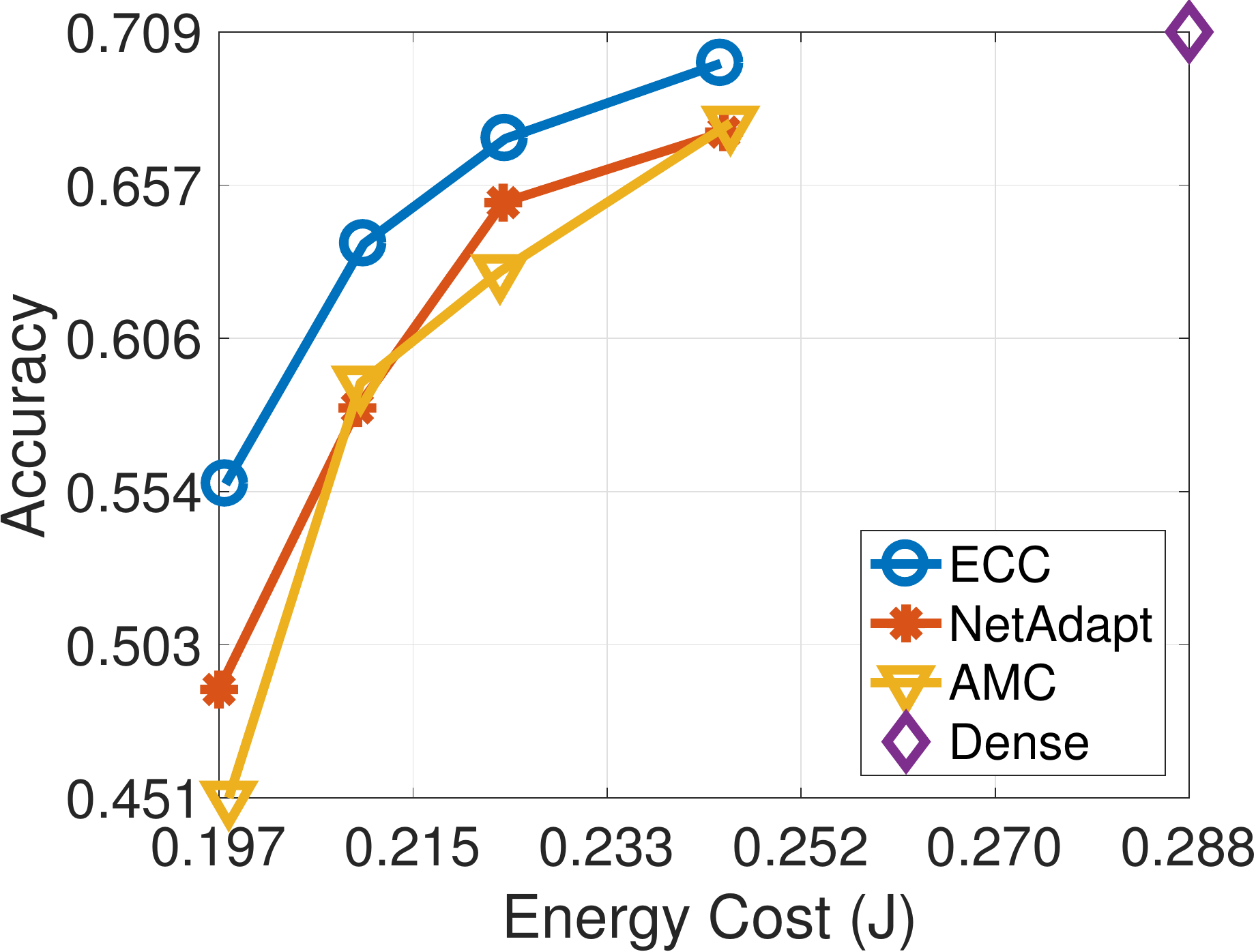}
		\label{fig:mb:1}
	}
	\subfloat[\small{Jetson TX2.}]
	{
		\includegraphics[width=0.48\columnwidth]{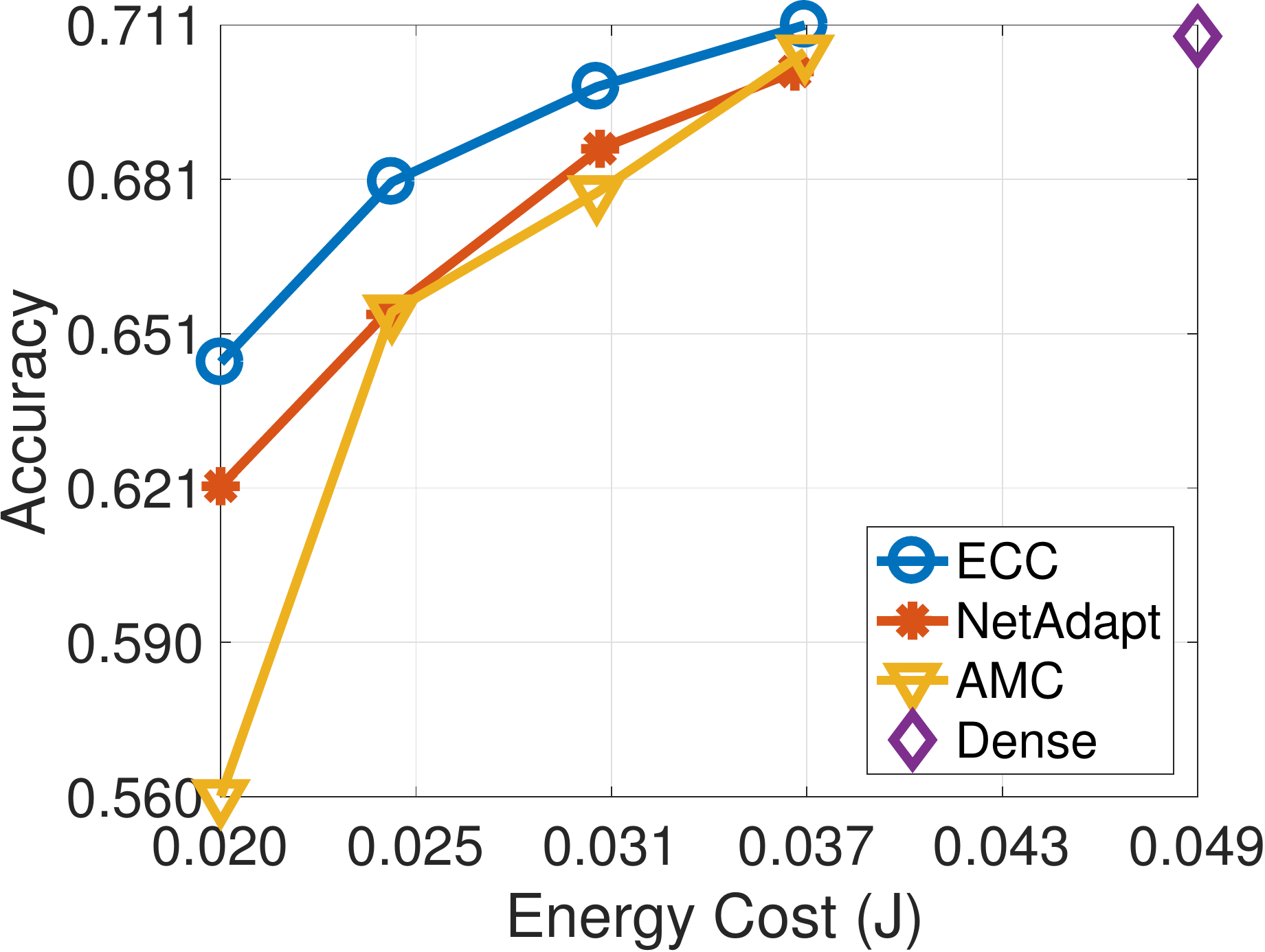}
		\label{fig:mb:3}
	}
	\vspace{-5pt}
	\caption{Top-1 accuracy of image classification on MobileNet@ImageNet \textbf{after} fine-tuning.}
	\label{fig:mb}
\end{figure}

\paragraph{MobileNet} In~\Fig{fig:mb}, we show the validation accuracies of compressed MobileNet under different energy budgets, and the energy cost is shown by joule (J). We set four different energy budgets in descending order. The dense MobileNet model has a top-1 accuracy of 0.709. The energy cost of the dense model is 0.2877~J on GTX 1080 Ti and 0.0487~J on Jetson TX2.

\vspace{-5pt}


\Fig{fig:mb} shows the top-1 accuracy v.s. energy comparisons (after fine-tuning) across the three methods. The results before fine-tuning is included in the supplementary material. ECC achieves higher accuracy than NetAdapt and AMC. For instance, on Jetson TX2 under the same 0.0247~J energy budget, ECC achieves 2.6\% higher accuracy compared to NetAdapt. Compared to the dense model, ECC achieves 37\% energy savings with $<1\%$ accuracy loss on Jetson TX2. AMC has similar performance with NetAdapt when the energy budget is not too small.

The accuracy improvements of ECC over NetAdapt and AMC are more significant under lower energy budgets. This suggests that under tight energy budget, searching for the optimal per-layer sparsity combinations becomes difficult, whereas ECC, via its optimization process, is able to identify better layer sparsities than search-based approaches.


\begin{figure}[htbp]
	\centering
	\subfloat[\small{GTX 1080~Ti.}]
	{
		\includegraphics[width=0.48\columnwidth]{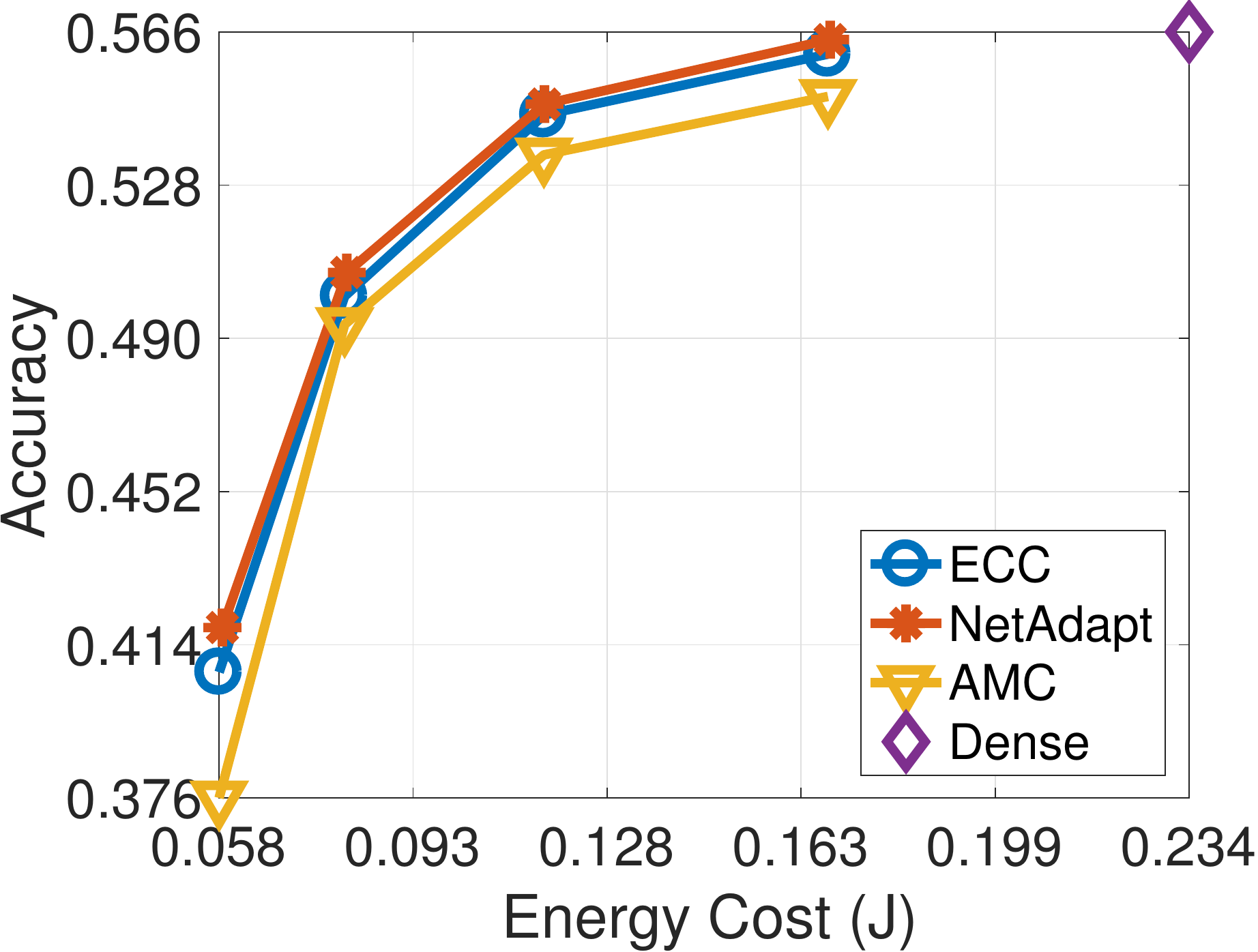}
		\label{fig:alex:1}
	}
	\subfloat[\small{Jetson TX2.}]
	{
		\includegraphics[width=0.48\columnwidth]{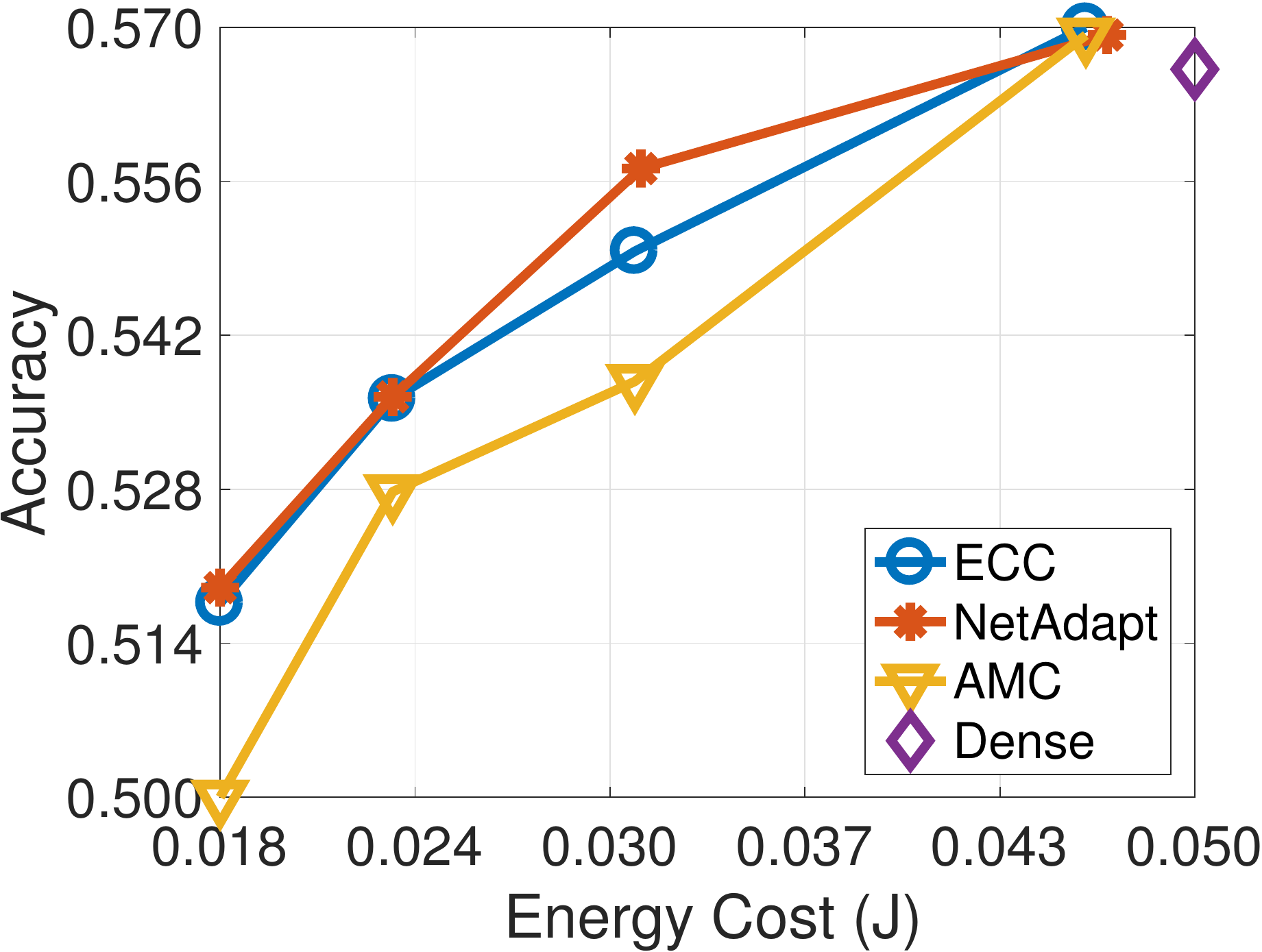}
		\label{fig:alex:3}
	}
	\vspace{-5pt}
	\caption{Top-1 accuracy image classification on AlexNet@ImageNet \textbf{after} fine-tuning.}
	\label{fig:alex}
	\vspace{-15pt}
\end{figure}


\begin{table*}[htp]
	\Huge
	\caption{Segmentation accuracy (averaged IoU) comparison on the Cityscapes dataset.}
	\label{tab:segment}
	\centering
	\renewcommand*{\arraystretch}{1.1}
	\renewcommand*{\tabcolsep}{12pt}
	\resizebox{2\columnwidth}{!}
	{
		\begin{tabular}{l|ccccc|ccccc}
			\toprule[0.15em]
			\multirow{2}{*}{Methods \slash\ Energy Budget} & \multicolumn{5}{c|}{GTX 1080 Ti}                                                                             & \multicolumn{5}{c}{Jetson TX2}                                                                              \\ \cline{2-11} & 1.3843 J & \multicolumn{1}{c}{1.4451 J} & \multicolumn{1}{c}{1.5238 J} & \multicolumn{1}{c}{1.6519 J} & \multicolumn{1}{c|}{1.9708 J} & 0.8542 J                              & 0.9051 J & 0.9756 J & \multicolumn{1}{c}{1.0724 J} & \multicolumn{1}{c}{1.3213 J} \\ 
			
			\midrule[0.05em]
			Dense & - & - & - & - & 0.722 & - & - & - & - & 0.722 \\
			\textbf{ECC} & \multicolumn{1}{c}{\textbf{0.6713}} & \textbf{0.6830} & \textbf{0.7007} & \multicolumn{1}{c}{\textbf{0.7163}} & \multicolumn{1}{c|}{-} & \multicolumn{1}{c}{\textbf{0.6733}} & \textbf{0.6914} & \textbf{0.7017} & \textbf{0.7183}  & \multicolumn{1}{c}{-} \\
			
			NetAdapt \cite{yang2018netadapt} & \multicolumn{1}{c}{0.6361} & 0.6523 & 0.6865 & 0.7114 & \multicolumn{1}{c|}{-} & \multicolumn{1}{c}{0.6567} & 0.6708 & 0.6916 & 0.7084 & \multicolumn{1}{c}{-}\\
			
			AMC \cite{he2018amc} & 0.6340 & \multicolumn{1}{c}{0.6374} & \multicolumn{1}{c}{0.6749} & \multicolumn{1}{c}{0.6992} & \multicolumn{1}{c|}{-} & 0.6344                               & \multicolumn{1}{c}{0.6491} & \multicolumn{1}{c}{0.6685} & \multicolumn{1}{c}{0.6976}  & \multicolumn{1}{c}{-}  \\
			\bottomrule[0.15em]
		\end{tabular}
	}
		\vspace{-12pt}
\end{table*}

\paragraph{AlexNet} We obtain similar conclusions on AlexNet. The dense model has a 0.566 top-1 accuracy. The energy cost of the dense model is 0.2339 J on GTX~1080~Ti and 0.0498 J on Jetson TX2. 
\Fig{fig:alex} compares the three methods (after fine-tuning) on two platforms respectively.
The results before fine-tuning are included in the supplementary material.
Before fine-tuning, ECC outperforms NetAdapt, which however achieves similar or slightly better accuracy than ECC after fine-tuning. ECC consistently outperforms AMC before and after fine-tuning. Compared to dense models, ECC achieves {28}\% and {37}\% energy savings with $<0.6\%$ and $<1.7\%$ accuracy loss on GTX~1080~Ti and Jetson TX2, respectively.

Comparing the results on AlexNet (7 layers) and MobileNet (14 layers), we find that the advantage of ECC is more pronounced on deeper networks. This is because as the network becomes deeper the layer sparsity search space grows exponentially, which makes the search-based approaches such as NetAdapt and AMC less effective.

\begin{figure}[t]
	\centering
	\subfloat[\small{MobileNet (inverse) sparsity.}]
	{
		\includegraphics[width=0.48\linewidth]{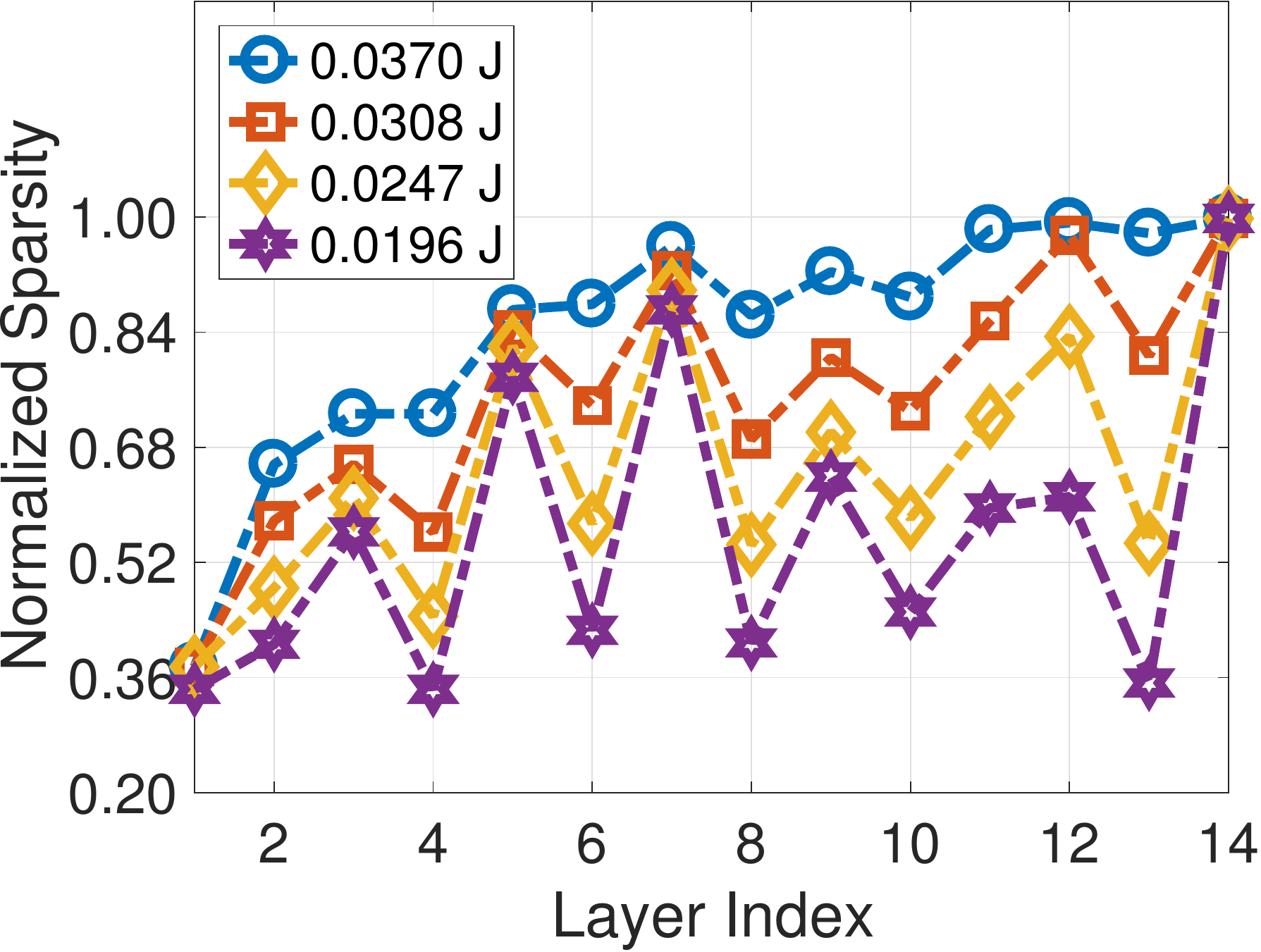}
		\label{fig:mb_ls}
	}
	\subfloat[\small{AlexNet (inverse) sparsity.}]
	{
		\includegraphics[width=0.48\linewidth]{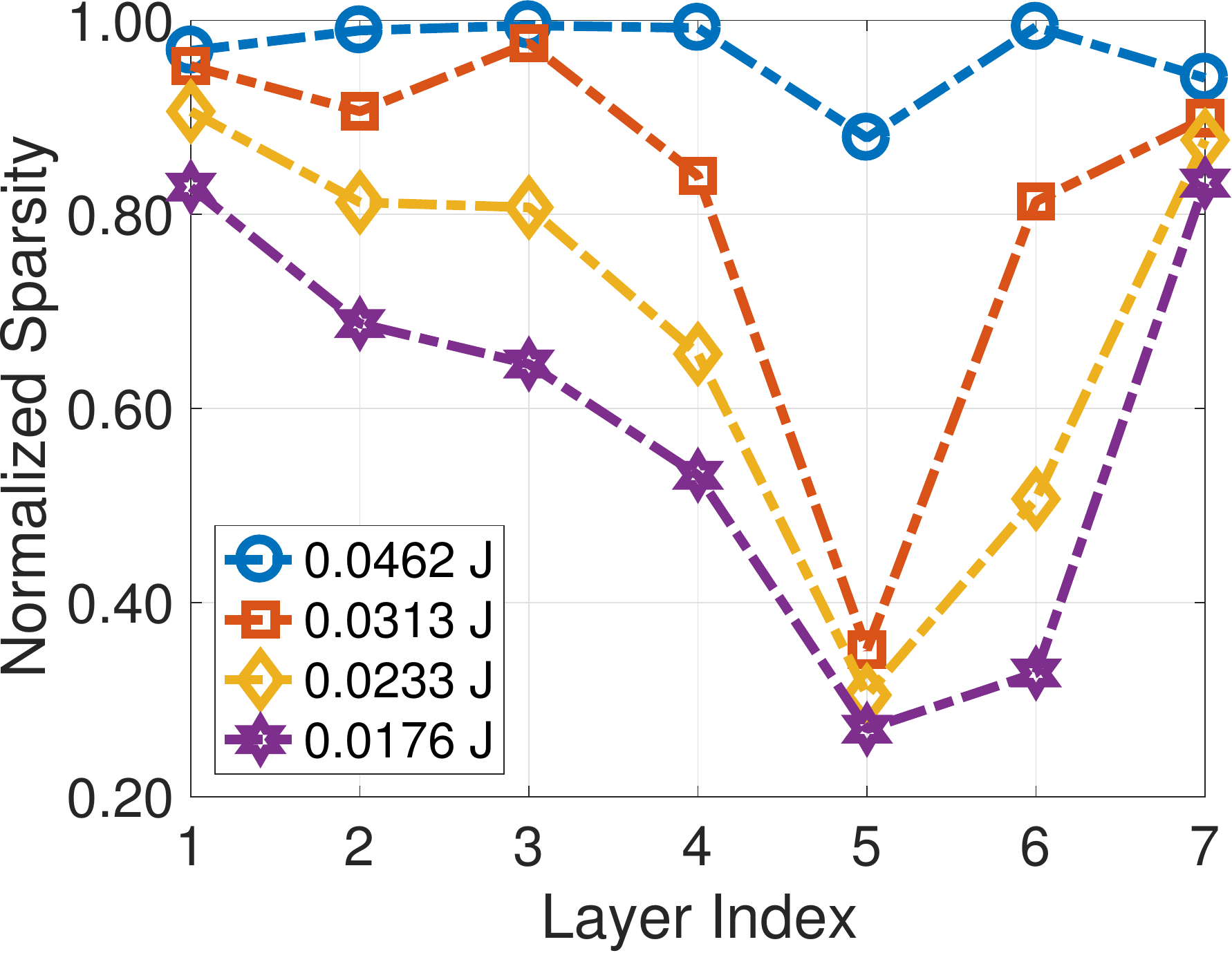}
		\label{fig:alex_ls}
	}
	\caption{Layer (inverse) sparsity after compressing on Jetson TX2.}
	\label{fig:sparsity}
	\vspace{-15pt}
\end{figure}

\paragraph{Sparsity Analysis}~\Fig{fig:mb_ls} and~\Fig{fig:alex_ls} show the normalized (inverse) sparsity (i.e. \#(nonzero channels)) of each layer in MobileNet and AlexNet respectively. Different colors represents different energy budgets used in~\Fig{fig:mb} and~\Fig{fig:alex}. We find that the $5^{th}$ layer in AlexNet is pruned heaviest. In AlexNet, that is the first fully connected layer which has the most number of weights; pruning it saves lots of energy. We also observe many ``spikes'' in MobileNet. Our intuition is that every two consecutive layers can be seen as a low rank factorization of a larger layer (ignoring the nonlinear activation between layers). The spikes may suggest that low rank structure could be efficient in saving energy.

\subsubsection{Cityscapes Segmentation}
\label{sec:eval:res:city}

Now we apply ECC to ERFNet~\cite{romera2018erfnet} for semantic segmentation. We use the well-established averaged Intersection-over-Union (IoU) metric, which is defined as $\text{TP}/(\text{FP}+\text{TP}+\text{FN})$ where TP, FP, and FN denote true positives, false positives, and false negatives, respectively. The training protocol is the same as the ImageNet experiments, except that the number of training iterations is 30,000 and the results are fine-tuned with 10,000 extra iterations. The dense model has an IoU of 0.722 and energy cost of 1.9708~J on GTX 1080 Ti and 1.3213~J on Jetson TX2.


\Tbl{tab:segment} compares the IoUs of the three compression techniques under different energy budgets. ECC consistently achieves the highest accuracy under the same energy budget. Similar to MobileNet, ERFNet is also a deep network with 51 layers, which leads to large layer sparsity search space that makes search-based approaches ineffective. Compared to the dense model, ECC reduces energy by 16\% and 19\% with $<0.6\%$ IoU loss on GTX 1080~Ti and TX2, respectively. 

\section{Conclusion}
\label{sec:conc}

Future computer vision applications will be increasingly operating on energy-constrained platforms such as mobile robots, AR headsets, and ubiquitous sensor nodes. To accelerate the penetration of mobile computer vision, this paper proposes ECC, a framework that compresses DNN to meet a given energy budget while maximizing accuracy. We show that DNN compression can be formulated as a constrained optimization problem, which can be efficiently solved using gradient-based algorithms without many of the heuristics used in conventional DNN compressions. Although targeting energy as a case study in the paper, our framework is generally applicable to other resource constraints such as latency and model size. We hope that our work is a first step, not the final word, toward heuristics-free, optimization-based DNN improvements.
{\small
\bibliographystyle{ieee}
\bibliography{refs}
}


\clearpage
 \onecolumn
\section*{Supplementary Material}
\subsection*{More Experiment Results}

\begin{figure}[htbp]
	\centering
	\subfloat[\small{GTX 1080~Ti.}]
	{
		\includegraphics[width=0.48\columnwidth]{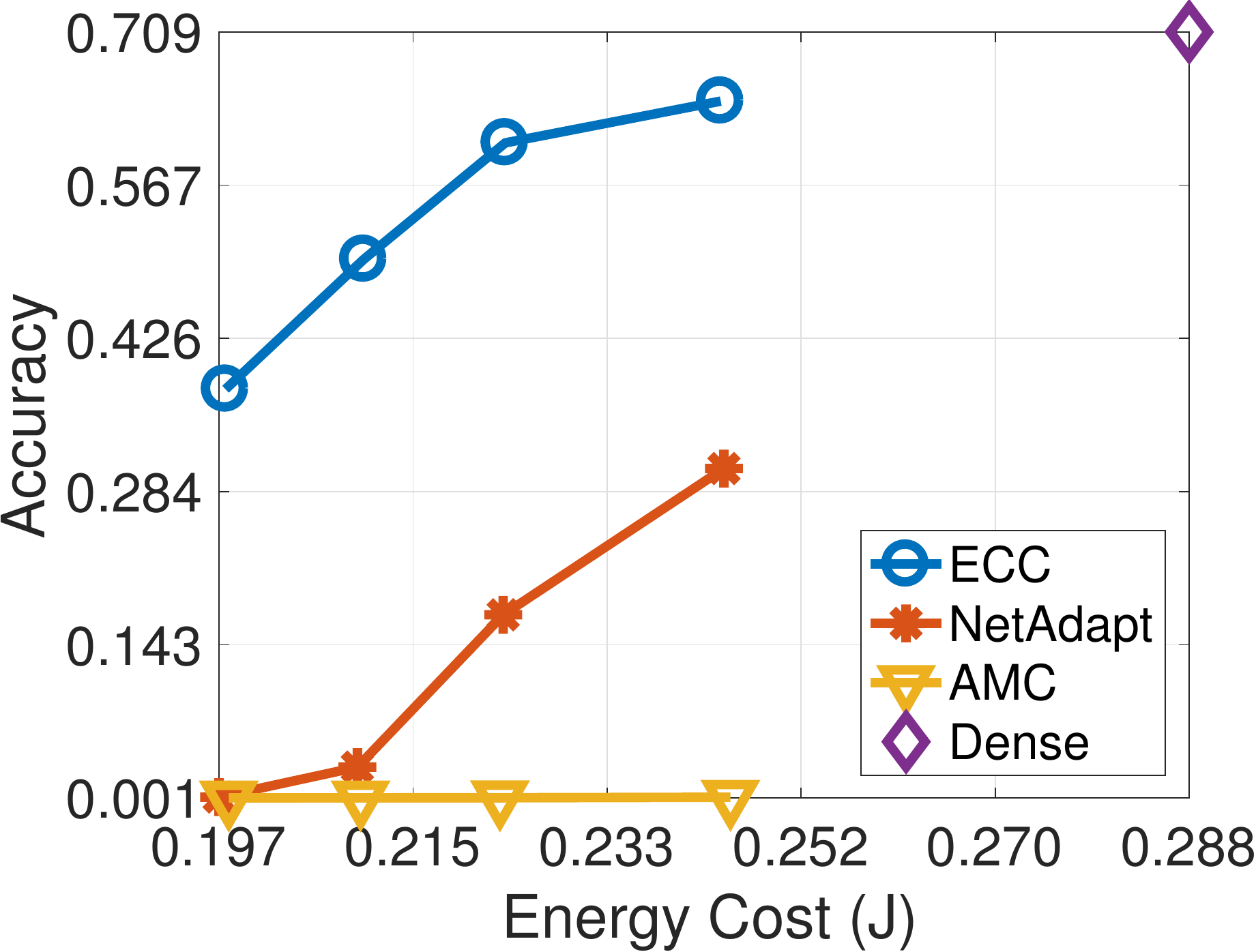}
		\label{fig:pftmb:1}
	}
	\subfloat[\small{Jetson TX2.}]
	{
		\includegraphics[width=0.48\columnwidth]{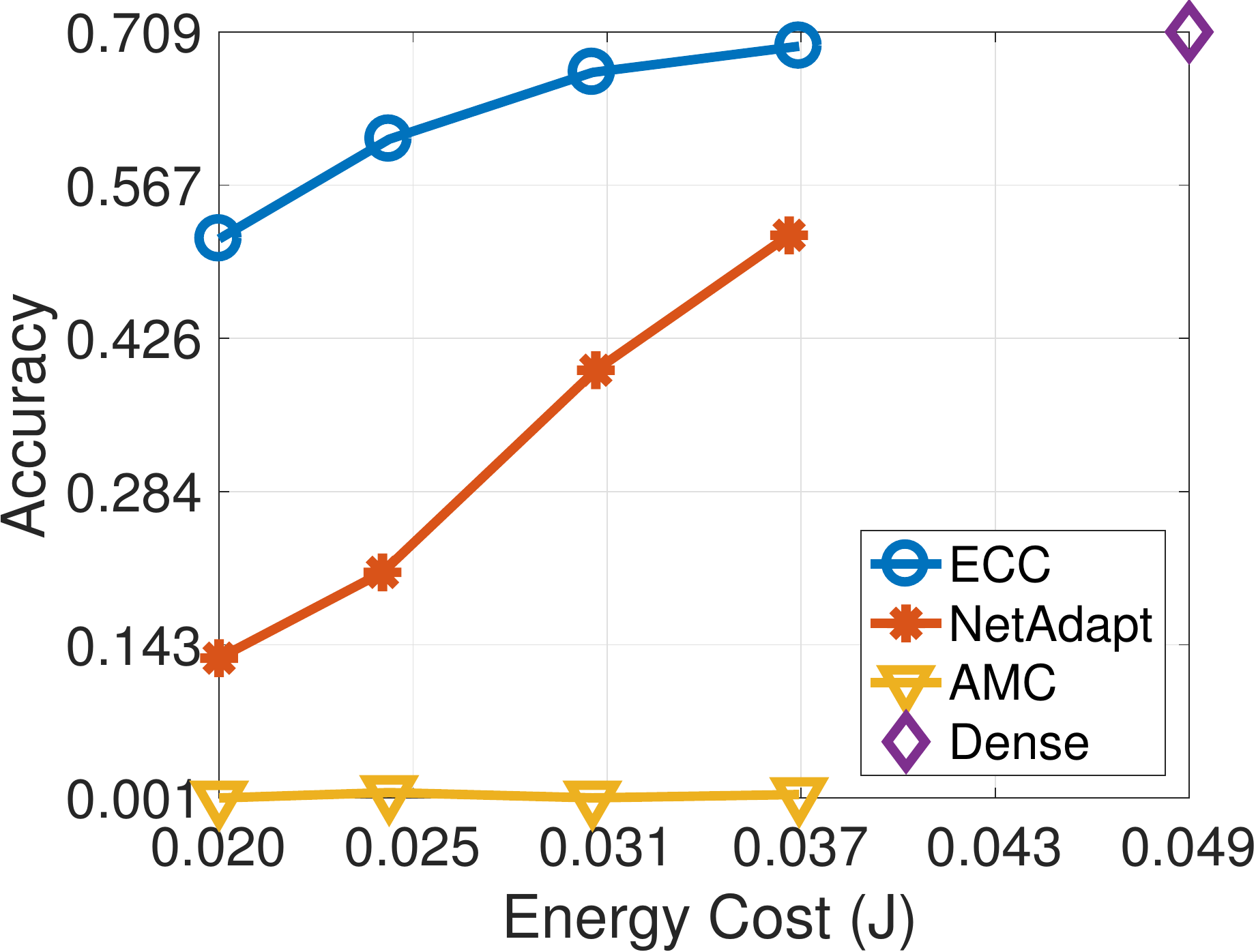}
		\label{fig:pftmb:3}
	}
	\caption{Top-1 accuracy of image classification on MobileNet@ImageNet \textbf{before} fine-tuning.}
	\label{fig:mbpft}
\end{figure}

\begin{figure}[htbp]
	\centering
	\subfloat[\small{GTX 1080~Ti.}]
	{
		\includegraphics[width=0.48\columnwidth]{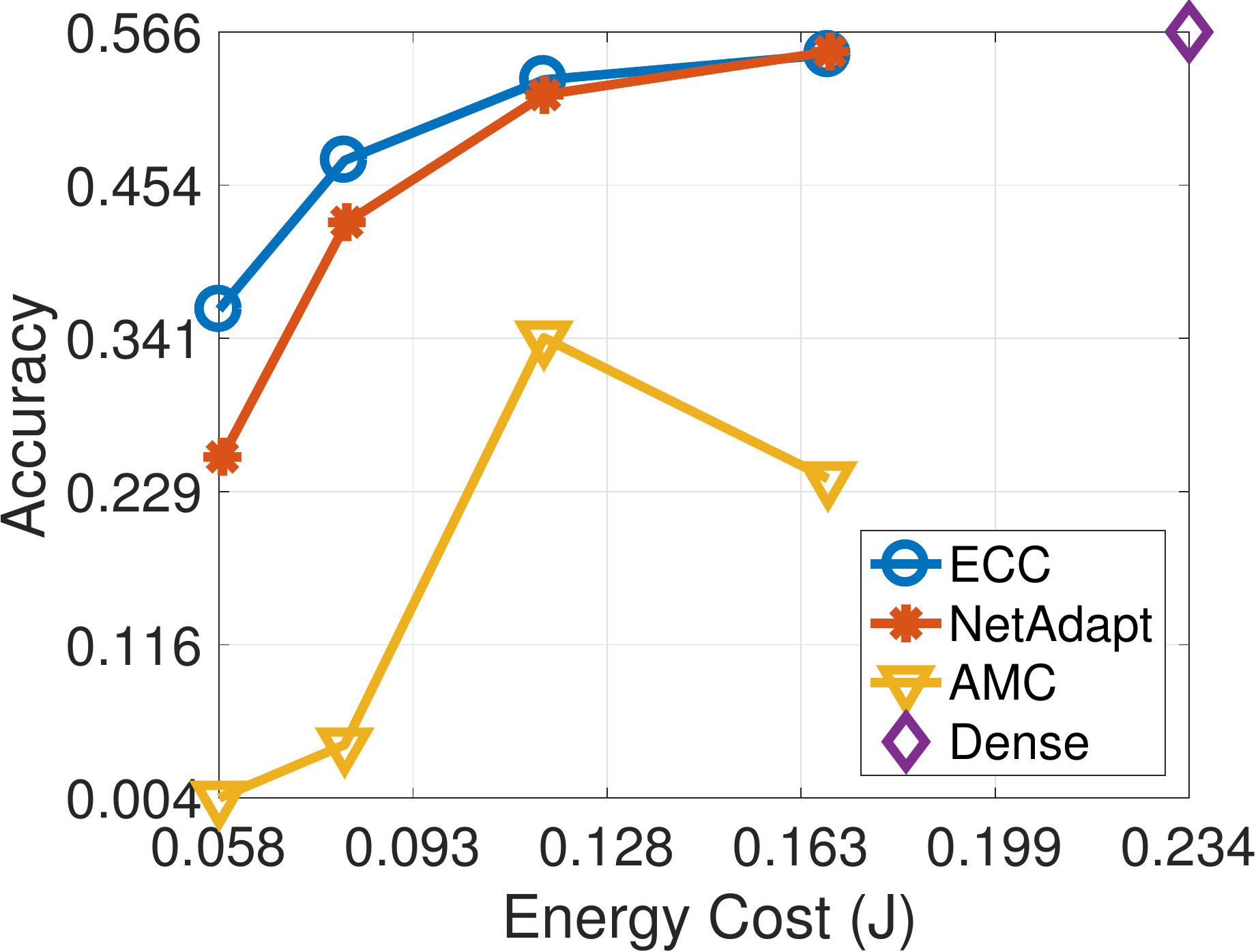}
		\label{fig:pftalex:1}
	}
	\subfloat[\small{Jetson TX2.}]
	{
		\includegraphics[width=0.48\columnwidth]{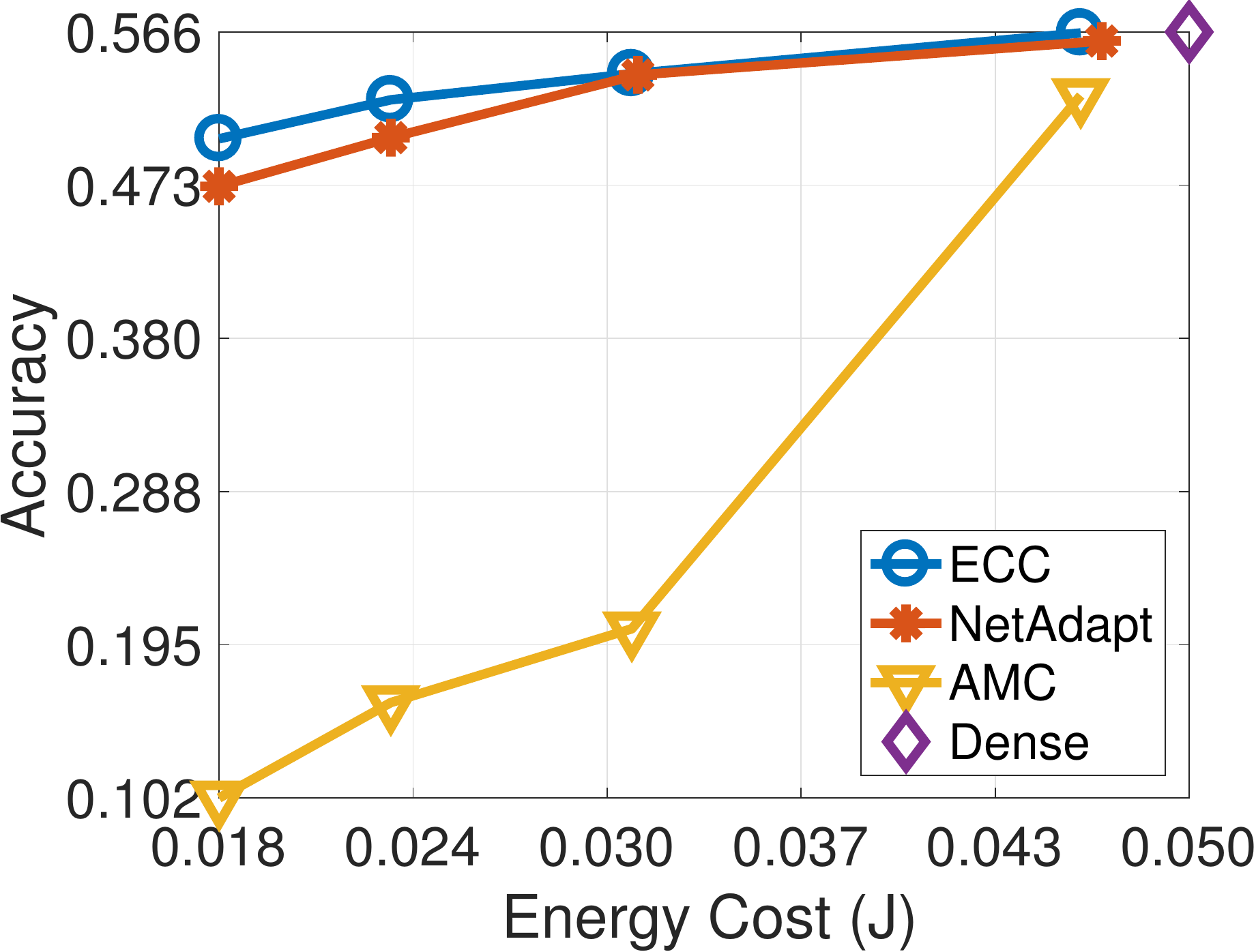}
		\label{fig:pftalex:3}
	}
	\caption{Top-1 accuracy of image classification on AlexNet@ImageNet \textbf{before} fine-tuning.}
	\label{fig:alexpft}
\end{figure}

\paragraph{ImageNet classification accuracy before fine-tuning}
The accuracy gains are significant before fine-tuning as shown in~\Fig{fig:mbpft} and ~\Fig{fig:alexpft}. For instance, on TX2 under the same 0.037~J energy budget, ECC achieves 17.5\% higher accuracy on MobileNet compared to NetAdapt. AMC has lower accuracies since it does not update DNN parameters in the RL searching phase.
Overall, we find that ECC is insensitive to additional fine-tuning while both NetAdapt and AMC require extensive fine-tuning to improve accuracy. This is because ECC, through its constrained optimization process, inherently performs compression and fine-tuning simultaneously.

\begin{figure}[htbp]
	\centering
	\subfloat[\small{AlexNet on GTX 1080~Ti.}]
	{
		\includegraphics[width=0.48\linewidth]{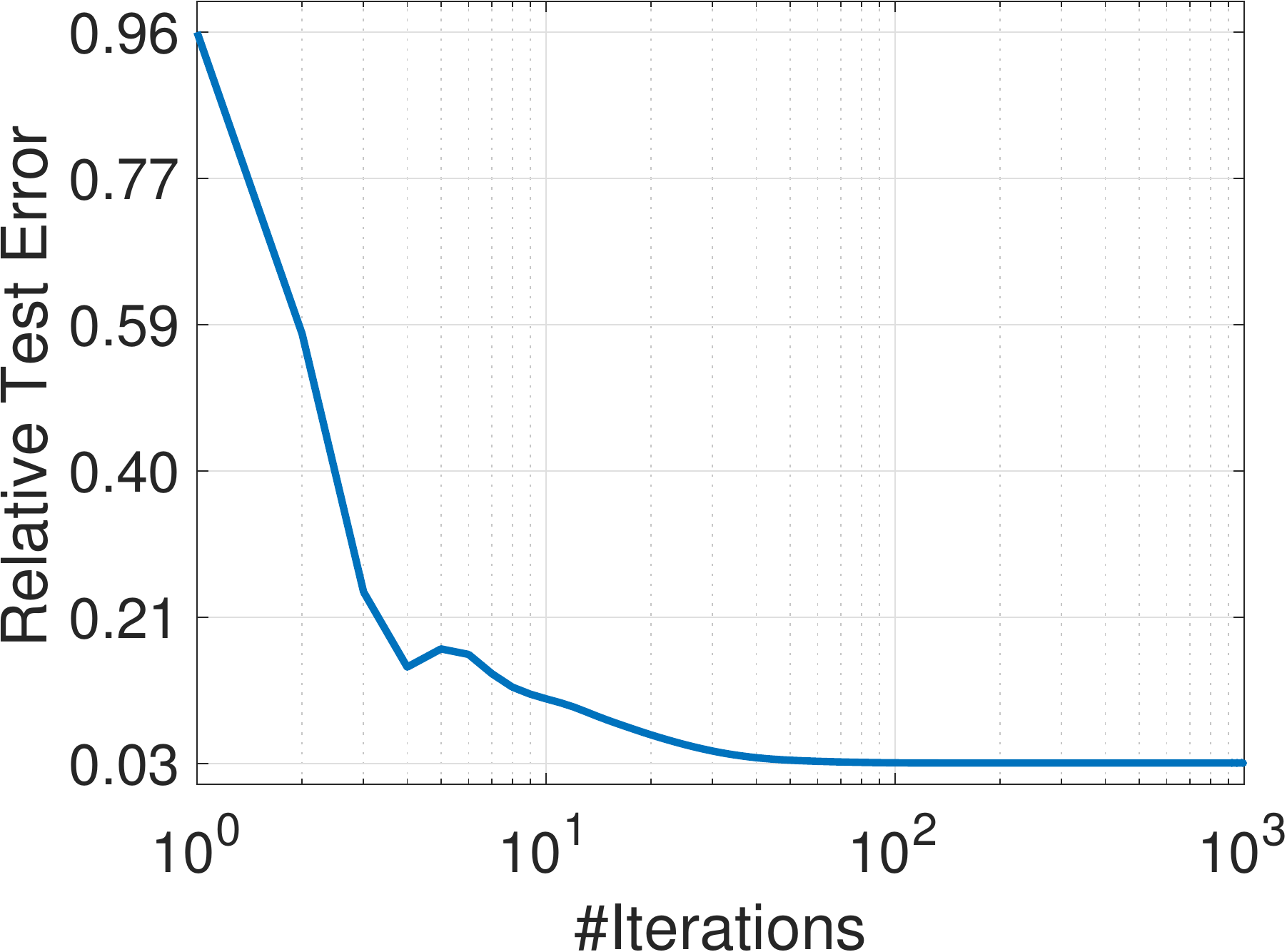}
	}
	\subfloat[\small{AlexNet on TX2.}]
	{
		\includegraphics[width=0.48\linewidth]{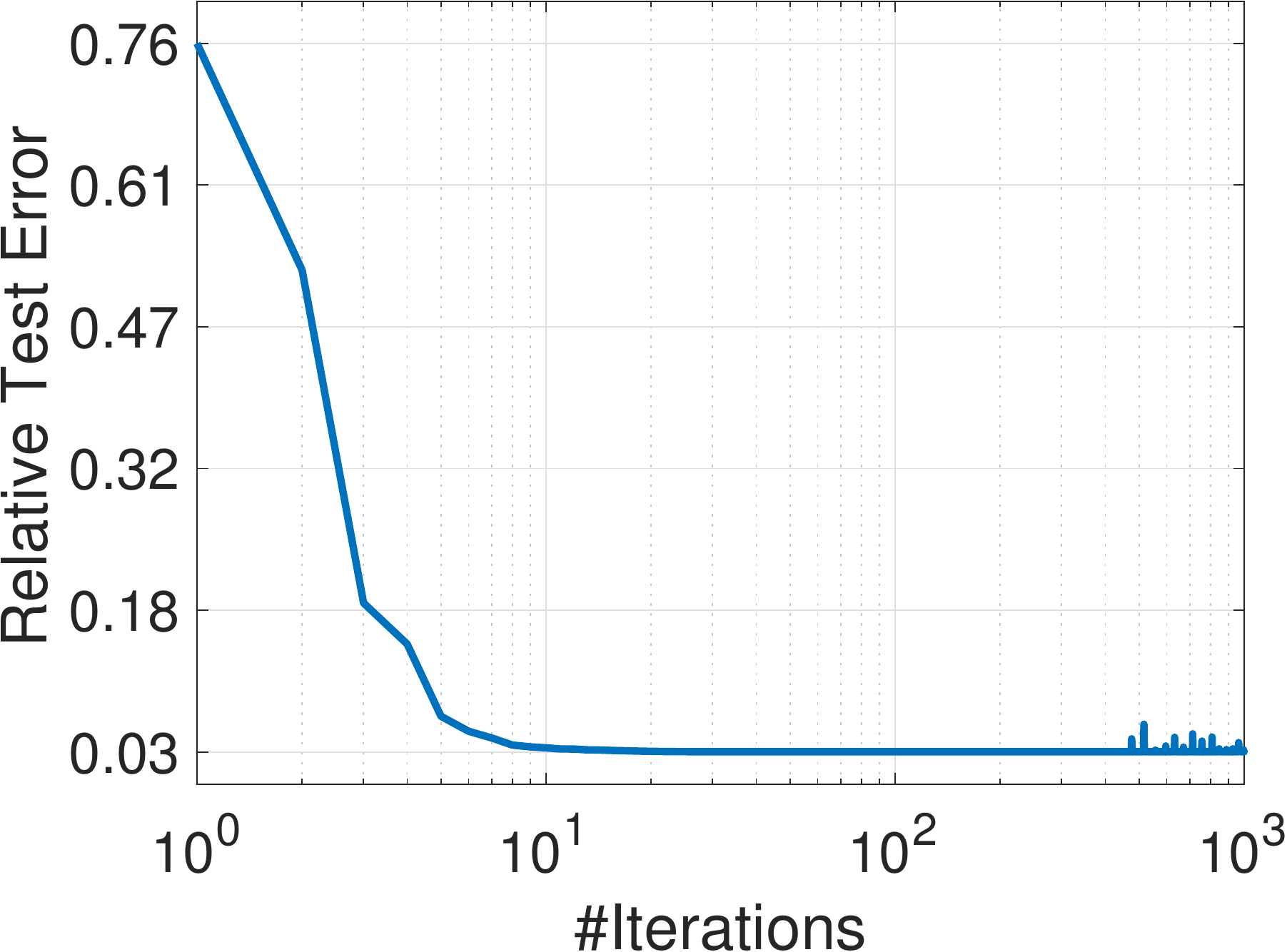}
	}
	\\
	\subfloat[\small{ERFNet on GTX 1080~Ti.}]
	{
		\includegraphics[width=0.48\linewidth]{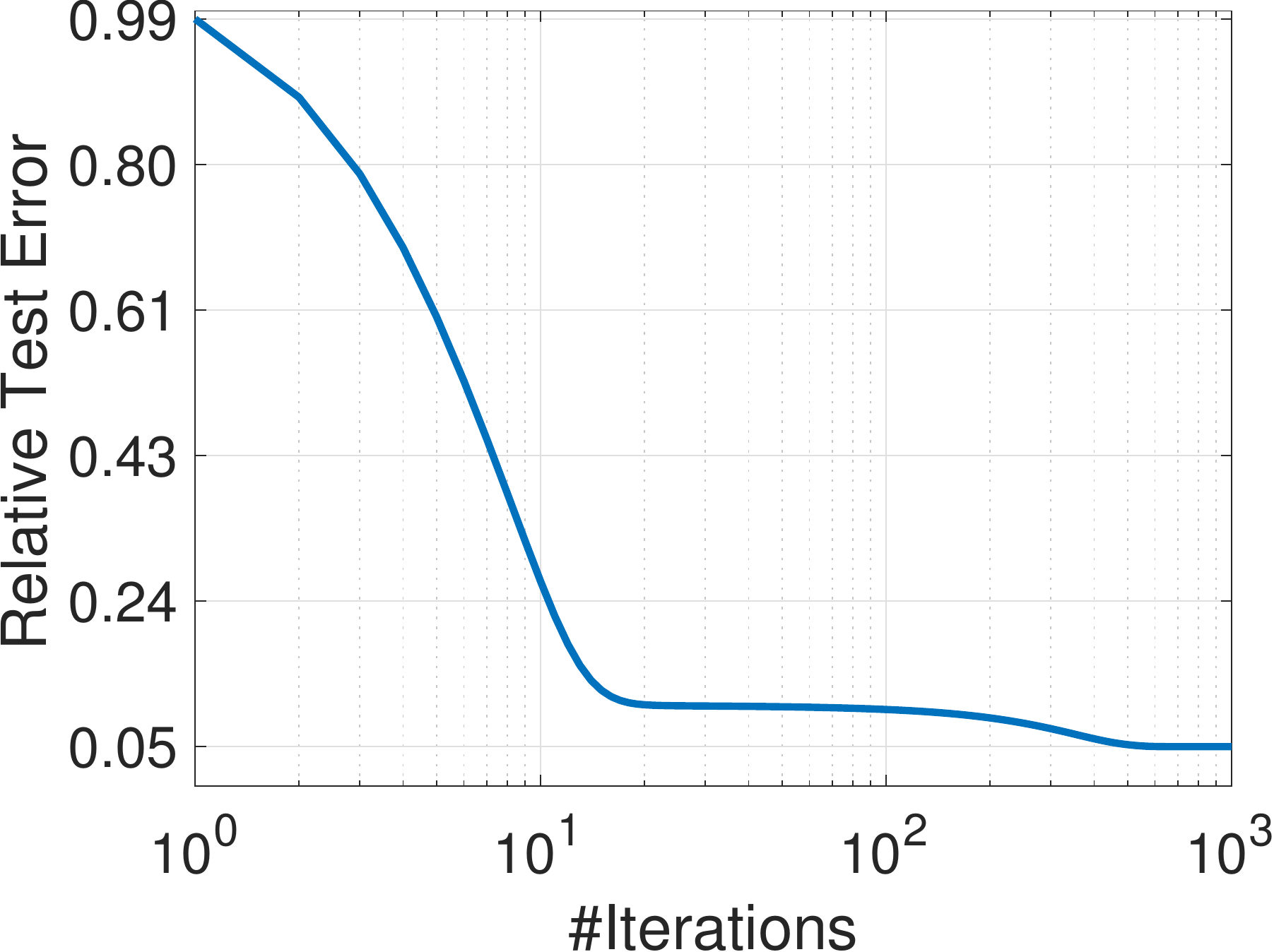}
	}
	\subfloat[\small{ERFNet on TX2.}]
	{
		\includegraphics[width=0.48\linewidth]{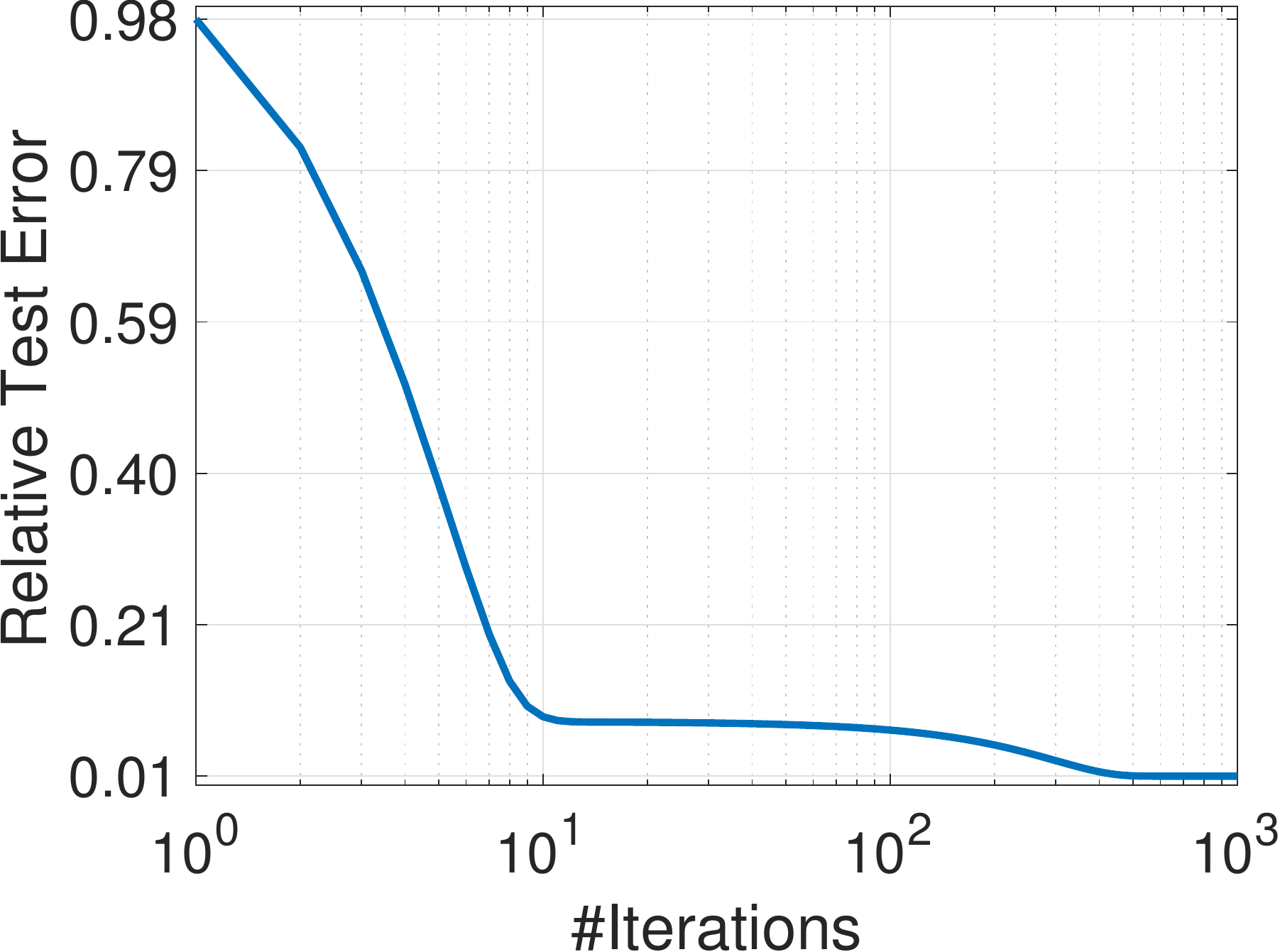}
	}
	\caption{Relative test error of energy prediction using the proposed bilinear model.}
	\label{fig:energy_supp1}
\end{figure}

\begin{figure}[htbp]
	\centering
	\subfloat[\small{AlexNet on GTX 1080~Ti.}]
	{
		\includegraphics[width=0.48\linewidth]{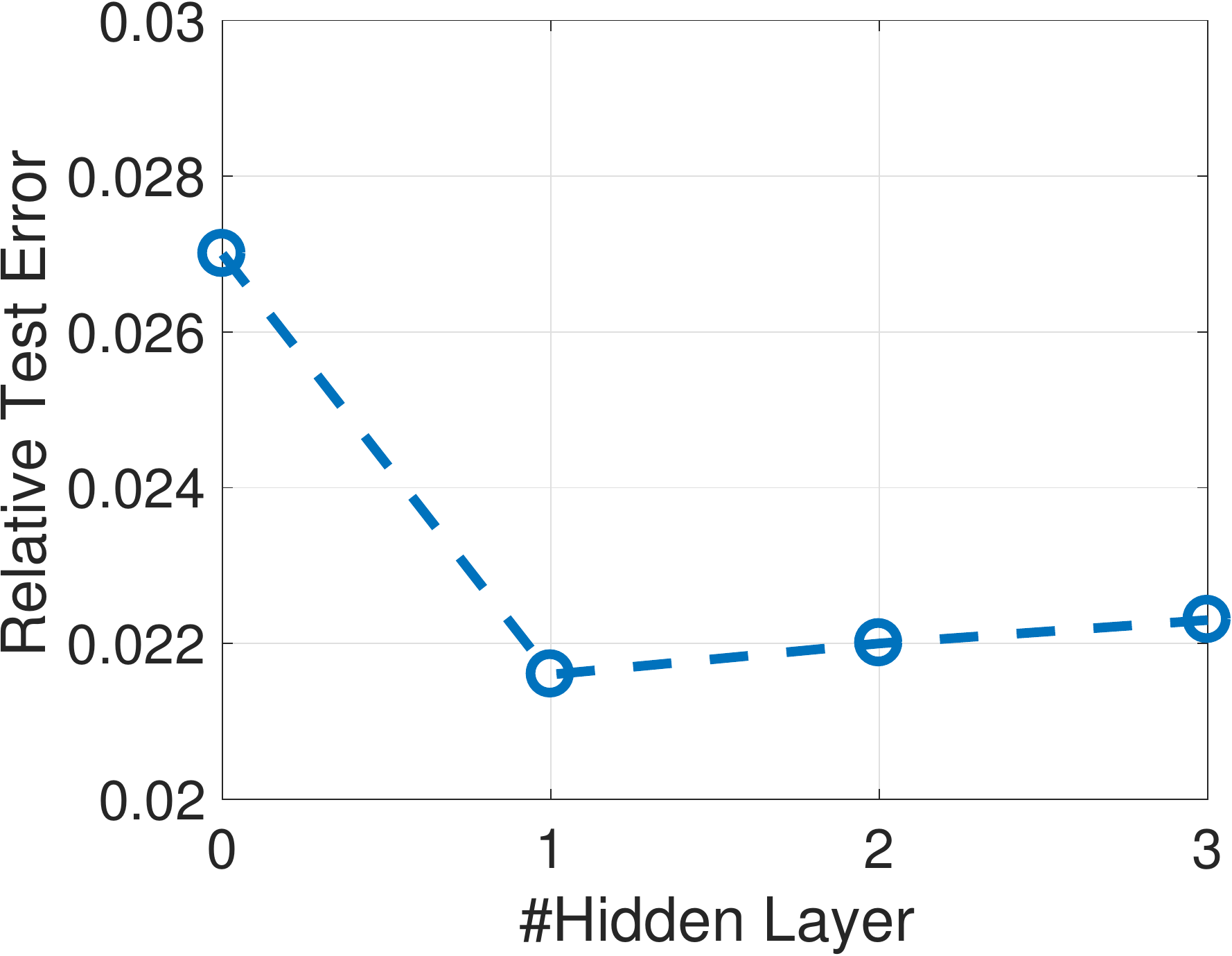}
	}
	\subfloat[\small{AlexNet on TX2.}]
	{
		\includegraphics[width=0.48\linewidth]{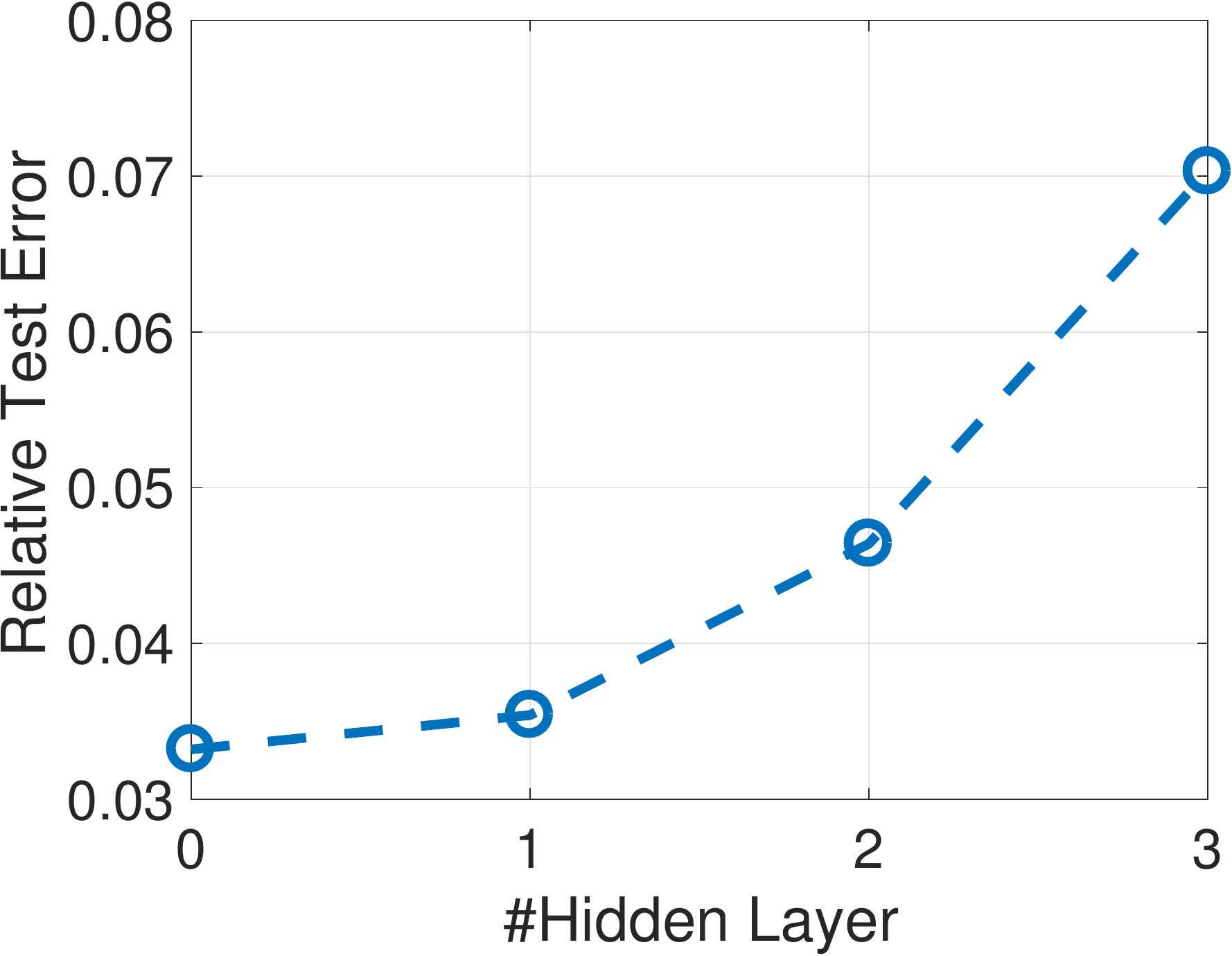}
	}
	\\
	\subfloat[\small{ERFNet on GTX 1080~Ti.}]
	{
		\includegraphics[width=0.48\linewidth]{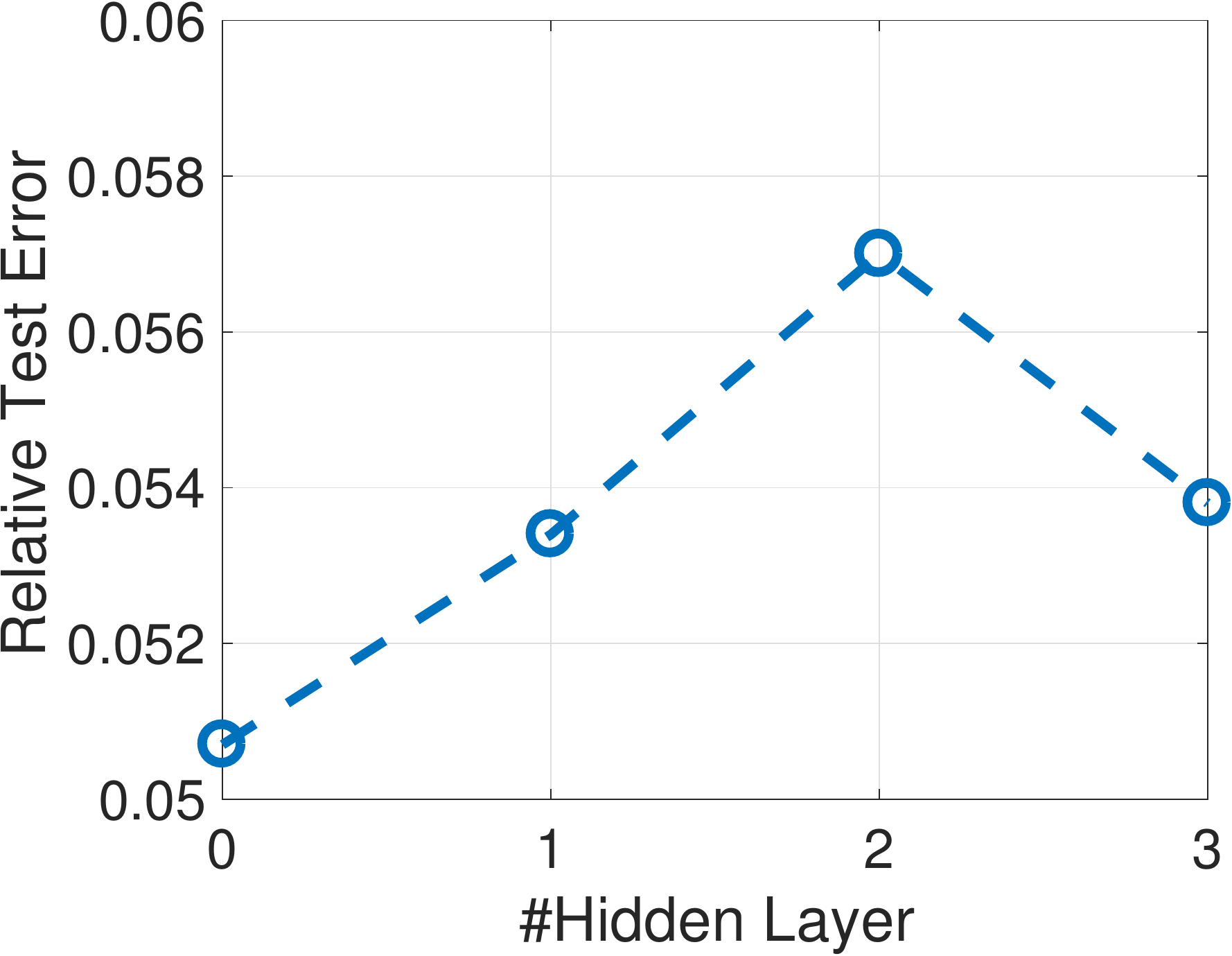}
	}
	\subfloat[\small{ERFNet on TX2.}]
	{
		\includegraphics[width=0.48\linewidth]{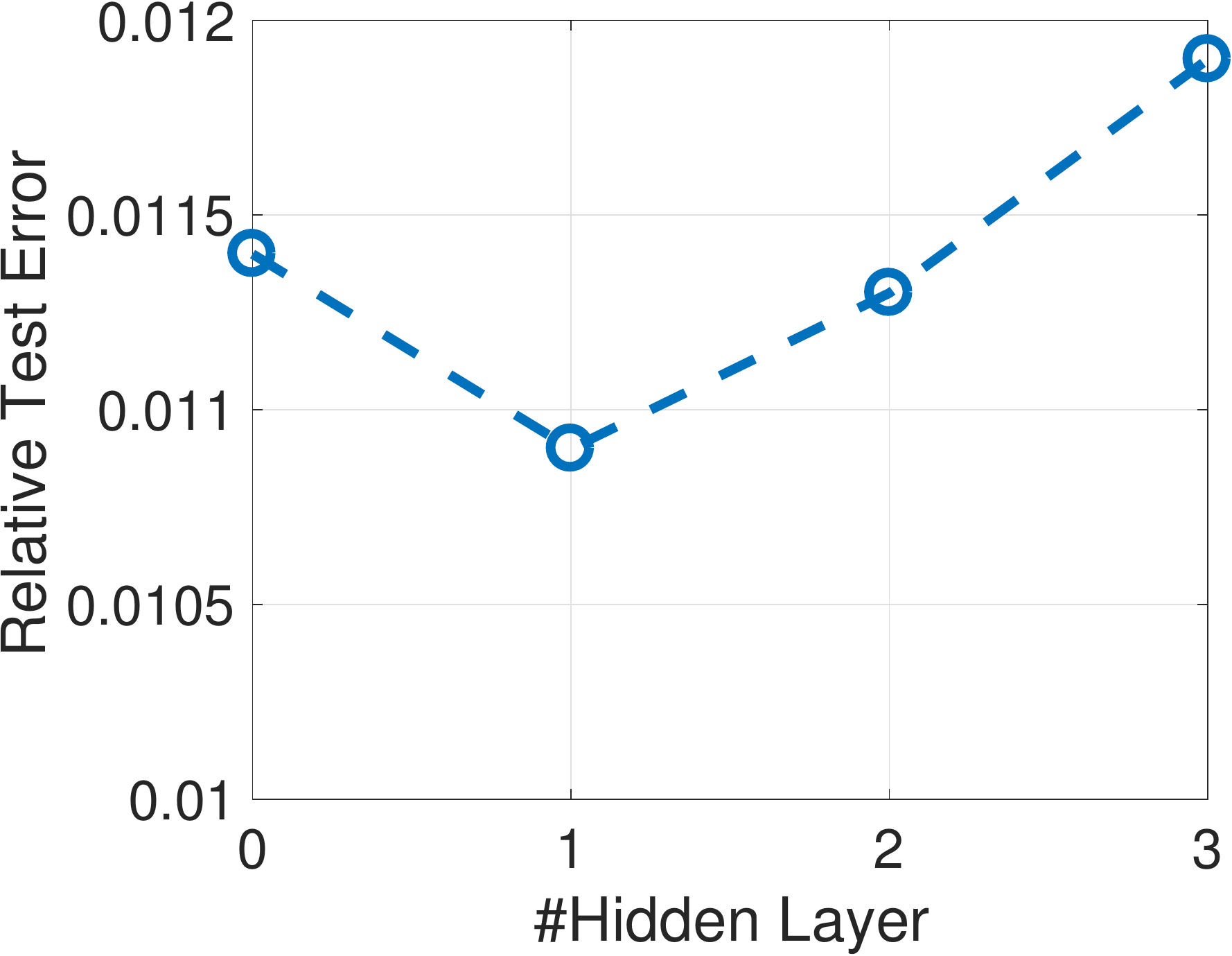}
	}
	\caption{Relative test error of energy prediction using an MLP model with different hidden layers.}
	\label{fig:energy_supp2}
\end{figure}

\paragraph{Energy model prediction errors on other networks}
Figures~\ref{fig:energy_supp1} and \ref{fig:energy_supp2} show the results of the energy prediction models as in Section~\ref{sec:eval:epred}.

\begin{figure}[ht]
	\centering
	\subfloat[\small{MobileNet on GTX 1080~Ti.}]
	{
		\includegraphics[width=0.48\linewidth]{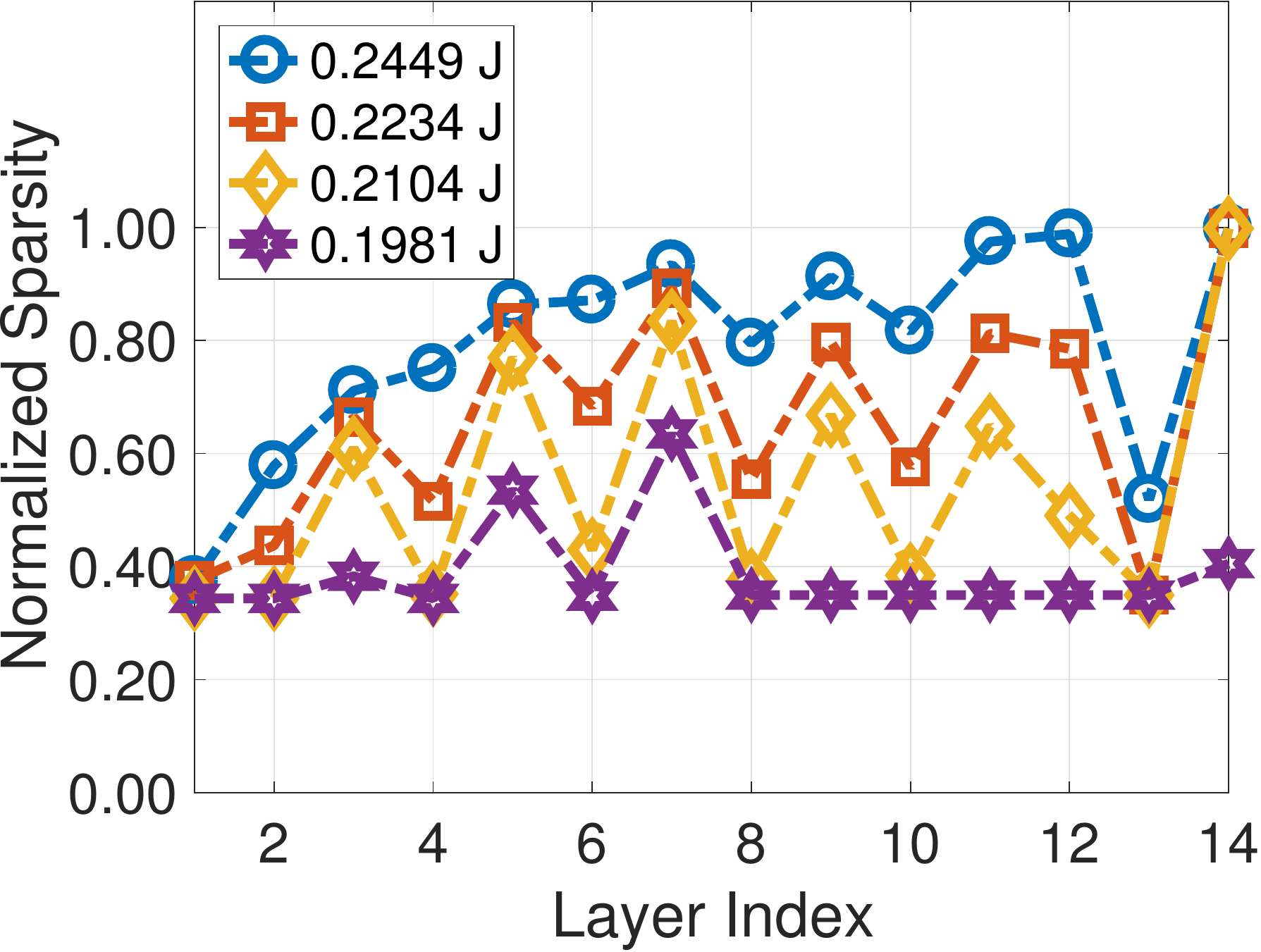}
	}
	\subfloat[\small{AlexNet on GTX 1080~Ti.}]
	{
		\includegraphics[width=0.48\linewidth]{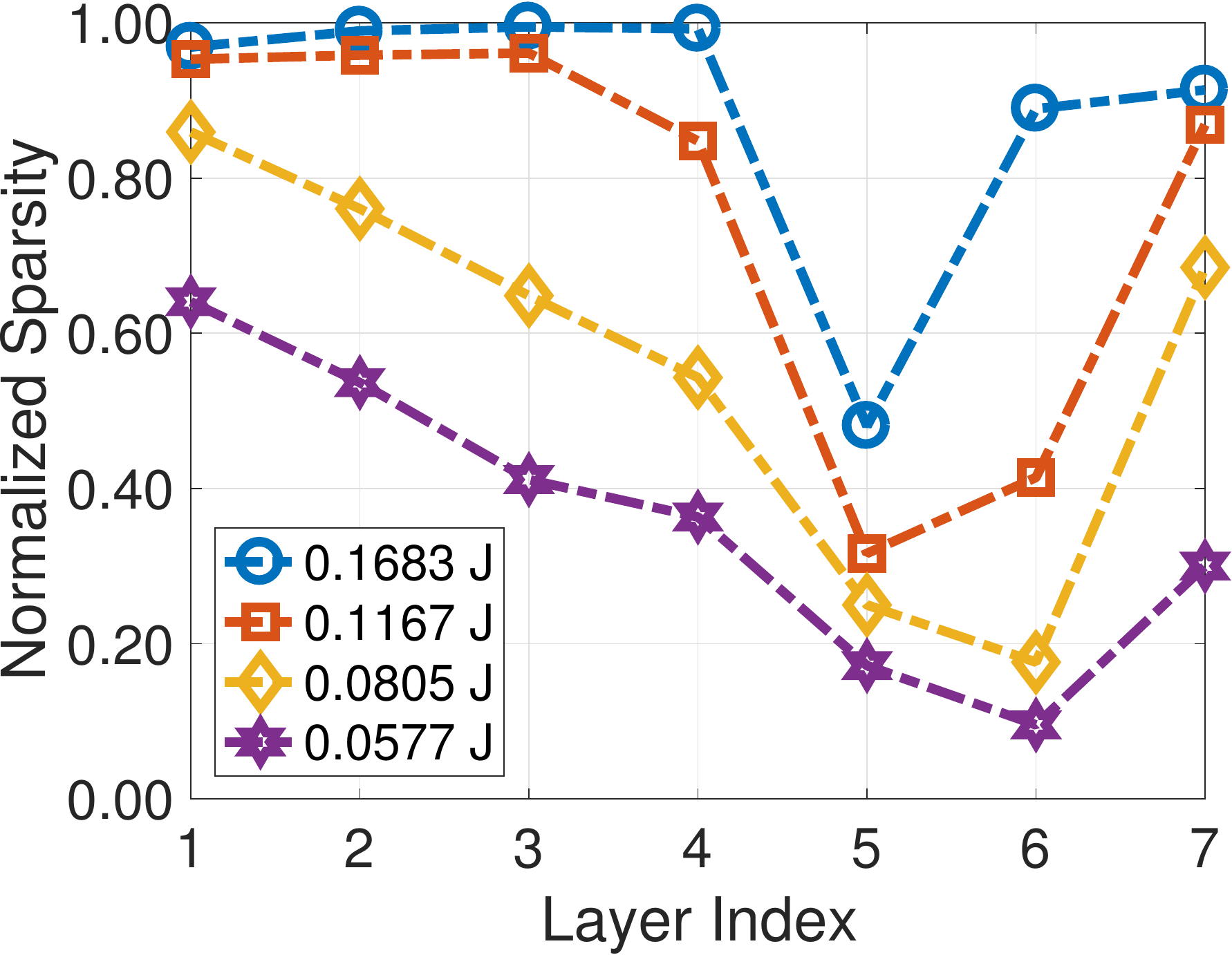}
	}
	\\
	\subfloat[\small{ERFNet on TX2.}]
	{
		\includegraphics[width=0.48\linewidth]{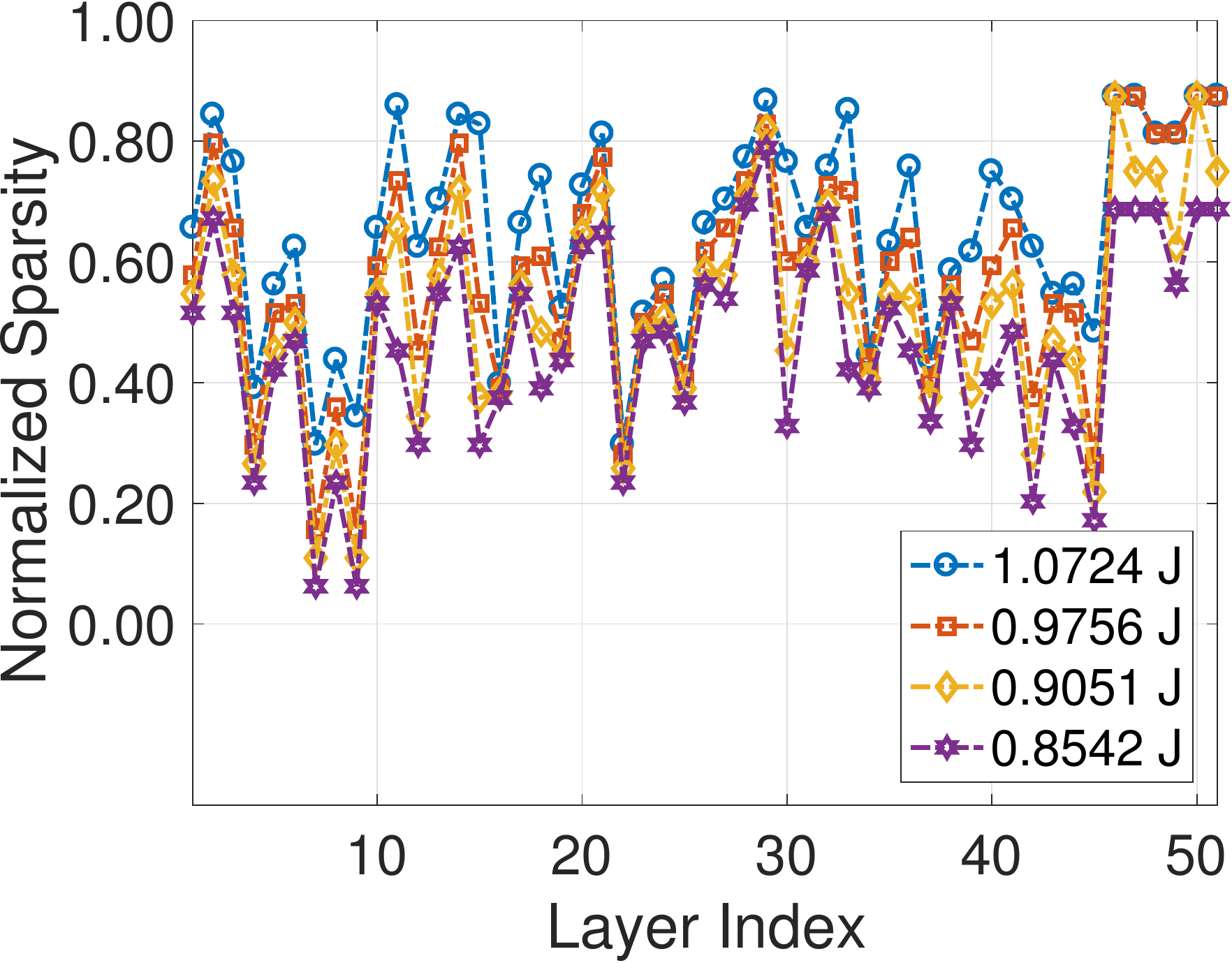}
	}
	\subfloat[\small{ERFNet on GTX 1080~Ti.}]
	{
		\includegraphics[width=0.48\linewidth]{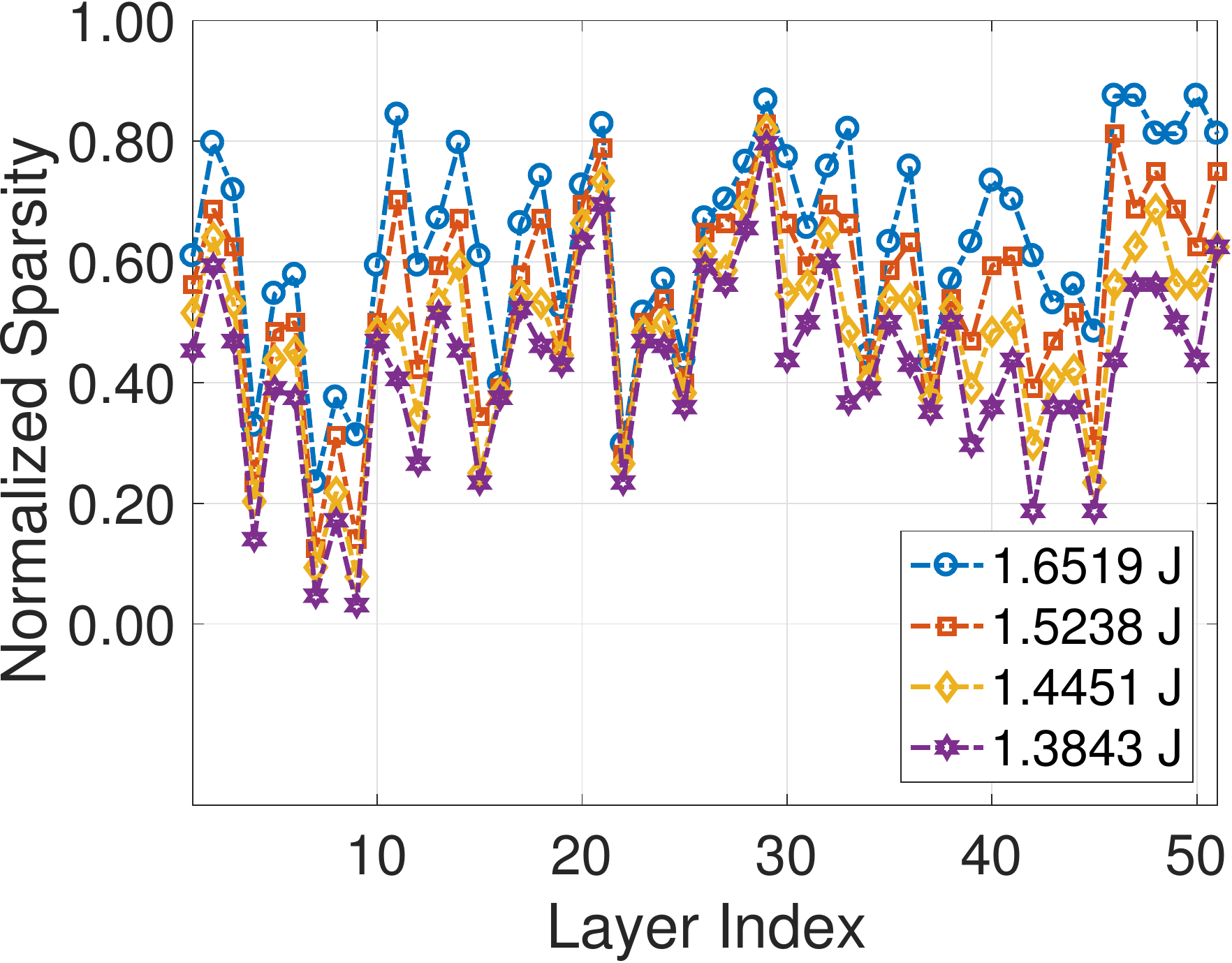}
	}
	\caption{Layer (inverse) sparsity after compressing.}
	\label{fig:ls_supp}
\end{figure}

\paragraph{Layer (inverse) sparsity on other networks}
Figures~\ref{fig:ls_supp} shows the layer (inverse) sparsity of the complementary compressed models of \Fig{fig:sparsity}. For MobileNet on GTX 1080 Ti, the lower bound of $s^{(u)}$ is set to be $0.35 c^{(u)}$.

\end{document}